%% file: kzsfd.tex
\documentclass[lettersize,journal]{IEEEtran}
\usepackage{amsmath}
\usepackage{amsmath, bm}
\usepackage{amsmath, amsfonts}
\usepackage{algorithmic}
\usepackage{algorithm}
\usepackage{array}
\usepackage[caption=false,font=normalsize,labelfont=sf,textfont=sf]{subfig}
\usepackage{textcomp}
\usepackage{stfloats}
\usepackage{url}
\usepackage{verbatim}
\usepackage{graphicx}
\usepackage{cite}
\usepackage{times}
\usepackage{epsfig}
\usepackage{amssymb}
\usepackage{microtype}

\usepackage[table,xcdraw]{xcolor}
\usepackage{caption}
\usepackage{placeins}
\usepackage{color, colortbl}
\usepackage{enumitem}
\usepackage{tabularx}
\usepackage{xstring}
\usepackage{multirow}
\usepackage{xspace}
\usepackage{booktabs}

\begin{document}

\title{Synthesizing Knowledge-enhanced Features for Real-world Zero-shot Food Detection}

\author{Pengfei Zhou,
Weiqing Min,~\IEEEmembership{Senior Member,~IEEE,}
Jiajun Song,
Yang Zhang,
and~Shuqiang Jiang,~\IEEEmembership{Senior Member,~IEEE}
\thanks{Manuscript received July 25, 2023; revised Jan 6, 2024; accepted Jan 24, 2024. This work was supported by the National Natural Science Foundation of China under Grant 61972378, Grant 62125207, Grant U1936203, and Grant U19B2040, and was also sponsored by the CAAI-Huawei MindSpore Open Fund.}
\thanks{P. Zhou, W. Min, J. Song, Y. Zhang, and S. Jiang are with the Key Laboratory of Intelligent Information Processing, Institute of Computing Technology, Chinese Academy of Sciences, Beijing 100190, China, and also with the University of Chinese Academy of Sciences, Beijing 100049 China. W. Min and S. Jiang are also with the Institute of Intelligent Computing Technology, Chinese Academy of Sciences, Suzhou 215124, China. \protect} 
\thanks{E-mail: pengfei.zhou@vipl.ict.ac.cn, minweiqing@ict.ac.cn, \{jiajun.song, yang.zhang\}@vipl.ict.ac.cn, sqjiang@ict.ac.cn.}% <-this % stops a space
}

\markboth{IEEE TRANSACTIONS ON IMAGE PROCESSING,~Vol.~XX, No.~X, July~2023}%
{P. Zhou \MakeLowercase{\textit{et al.}}: Synthesizing Knowledge-enhanced Features for Real-world Zero-shot Food Detection}

\maketitle

\input{0_abstract.tex}


\input{1_introduction.tex}
\input{2_related_work.tex}
\input{3_method_def.tex}
\input{4_method_main.tex}
\input{5_dataset.tex}
\input{6_experiments.tex}
\input{7_conclusion.tex}

{\small
\bibliographystyle{IEEEtran}
\bibliography{kzsfd}
}

\end{document}

%% file: 0_abstract.tex
% DONE:
\begin{abstract}
    Food computing brings various perspectives to computer vision like vision-based food analysis for nutrition and health. As a fundamental task in food computing, food detection needs Zero-Shot Detection (ZSD) on novel unseen food objects to support real-world scenarios, such as intelligent kitchens and smart restaurants. Therefore, we first benchmark the task of Zero-Shot Food Detection (ZSFD) by introducing FOWA dataset with rich attribute annotations. Unlike ZSD, fine-grained problems in ZSFD like inter-class similarity make synthesized features inseparable. The complexity of food semantic attributes further makes it more difficult for current ZSD methods to distinguish various food categories. To address these problems, we propose a novel framework ZSFDet to tackle fine-grained problems by exploiting the interaction between complex attributes. Specifically, we model the correlation between food categories and attributes in ZSFDet by multi-source graphs to provide prior knowledge for distinguishing fine-grained features. Within ZSFDet, Knowledge-Enhanced Feature Synthesizer (KEFS) learns knowledge representation from multiple sources (e.g., ingredients correlation from knowledge graph) via the multi-source graph fusion. Conditioned on the fusion of semantic knowledge representation, the region feature diffusion model in KEFS can generate fine-grained features for training the effective zero-shot detector. Extensive evaluations demonstrate the superior performance of our method ZSFDet on FOWA and the widely-used food dataset UECFOOD-256, with significant improvements by 1.8\% and 3.7\% ZSD mAP compared with the strong baseline RRFS. Further experiments on PASCAL VOC and MS COCO prove that enhancement of the semantic knowledge can also improve the performance on general ZSD. Code and dataset are available at https://github.com/LanceZPF/KEFS.
\end{abstract}

%  semantic information enhance   

%% file: 1_introduction.tex
\section{Introduction}
\label{sec:intro}

Food computing, an interdisciplinary research area that combines machine learning and food science to understand food-related data, has received considerable attention in computer vision~\cite{survey, Gayathri-Healthy-Nature2021,wang2022learning}. Its importance is spotlighted by its ability to enable future nutrition and health management through the analysis of visual content from food images~\cite{willett2019food}. Academically, it enriches computer vision research by providing complex real-world scenarios and fine-grained issues for challenging advanced image processing algorithms~\cite{min2023large, marin2019recipe1m, Jiang-MSMVDFA-TIP2019}. Industrially, it revolutionizes the food industry with enhanced agricultural strategies, automatic food processing, and personalized dietary assessment~\cite{basso2020digital, khan2022machine, wang2022review}. Food detection serves as a fundamental technique in food computing, applying object detection paradigms to various real-world scenarios~\cite{aguilar2018grab}, such as automatic settlement~\cite{survey} and dietary assessment~\cite{Meyers-Im2Calories-ICCV2015,wang2022review}. 

However, current food detection models face significant challenges in real-world scenes like intelligent kitchens~\cite{damen2020epic} and smart restaurants~\cite{aguilar2018grab}, where novel food classes constantly emerge. The conventional food detection systems based on detectors trained on fixed classes have difficulty recognizing new, unseen food categories, leading to limited practical utility. To this end, we formally benchmark Zero-Shot Food Detection (ZSFD) based on Zero-Shot Detection (ZSD) to bridge this gap by enabling the detection of unseen food objects without requiring labeled new data, which are occasionally unavailable for real-world food applications.

\begin{figure}[t]
	\centering
  \includegraphics[width=8.2cm]{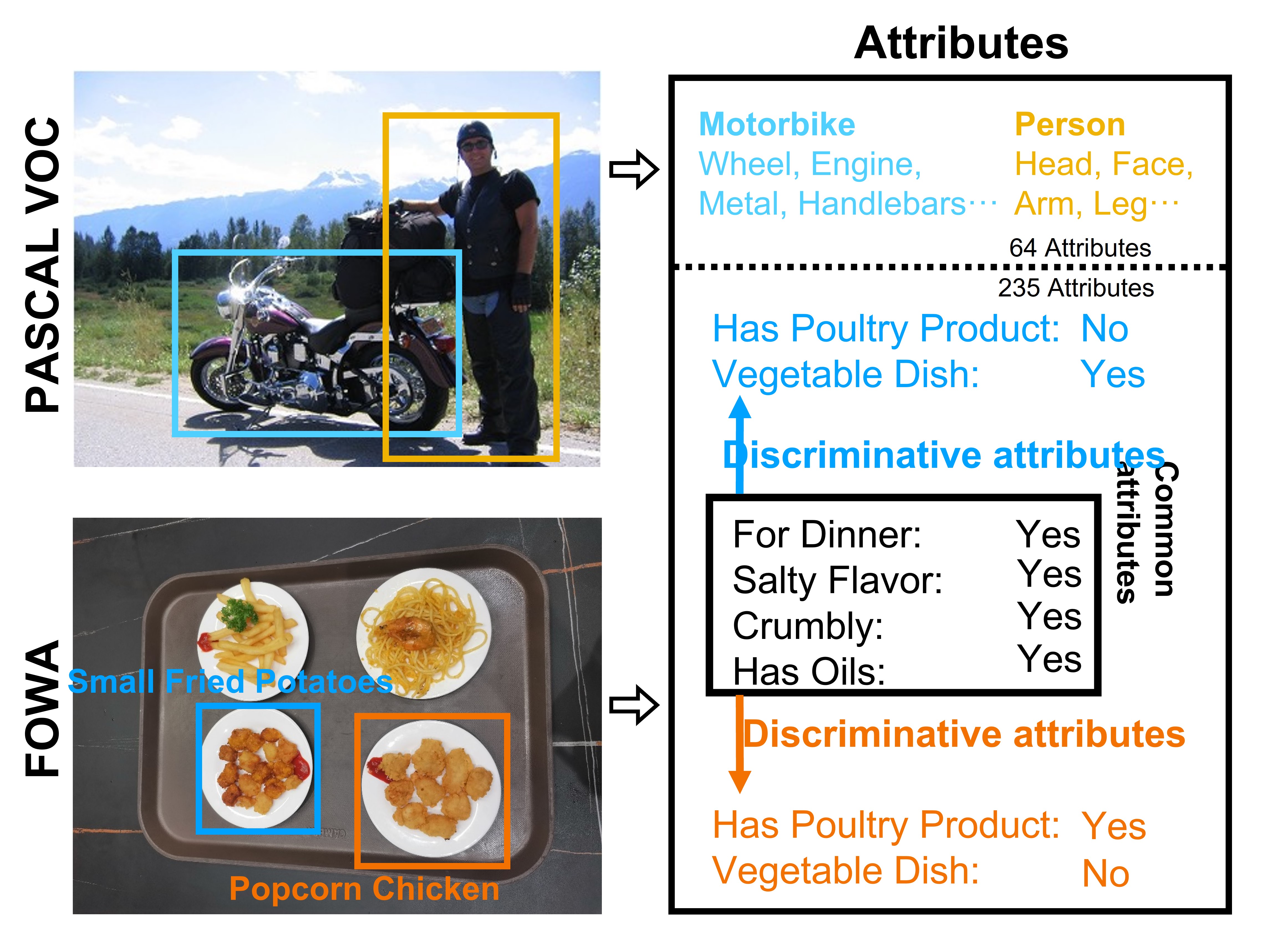}
  \caption{Description of problems on ZSFD. Food objects have inter-class similarity and attribute complexity.}
  % \vspace{-0.3cm}
  \label{fig:problem} 
\end{figure} 

To detect unseen objects “in the wild” scenarios, ZSD employs two major approaches: mapping-based methods \cite{bansal2018zero, berkan2018zero, yan2022semantics} and generation-based methods \cite{zhu2020don, hayat2020synthesizing, huang2022robust}. However, since the mapping-based models are trained only with seen features in training dataset, they are easily biased towards seen classes in Generalized Zero-Shot Detection (GZSD), where images contain objects from both seen classes and unseen classes. Recently, generation-based methods have emerged with better ZSD performance by addressing the bias problem based on generative models. Generative models such as Variational Auto-Encoder (VAE) and Generative Adversarial Networks (GAN) are used to synthesize unseen visual features from semantic vectors and transfer the zero-shot problem into supervised learning on both seen and synthesized unseen features. However, the performance of those methods is limited in ZSFD systems due to fine-grained problems like inter-class similarity and intra-class variability.

In the domain of ZSFD, generation-based methods require the synthesis of discriminative features. However, previous approaches meet difficulties in synthesizing distinguishable features for fine-grained food categories due to the problem of \textbf{inter-class similarity} and \textbf{attribute complexity}. As shown in Figure~\ref{fig:problem}, compared with common objects that have distinct patterns and certain semantic parts, \emph{Small Fried Potatoes} and \emph{Popcorn Chicken} have similar visual features due to the inter-class similarity. Furthermore, common objects have simple attributes strongly associated with their category, while the complex attributes of food objects make food categories more likely to be confused. For example, the \emph{Head} and \emph{Face} attributes of the \emph{People} can be strongly associated with a certain semantic part, which helps distinguish it from the \emph{Motorbike}. In contrast, the same attributes of flavor and mouthfeel further confuse those two visually similar categories of \emph{Small Fried Potatoes} and \emph{Popcorn Chicken}. Complex attributes require rich semantic information to describe, and existing semantic representations from word embeddings are insufficient to capture discriminative information about food attributes. 

Addressing these challenges necessitates a novel approach that leverages food domain knowledge, such as ingredient and hyperclass correlations. To this end, we introduce Zero-Shot Food Detector (ZSFDet), a new framework that utilizes domain knowledge from multiple sources, including ingredient correlation, hyperclass relationship and label co-occurrence probabilities modeled by knowledge, hyperclass and probability graphs, respectively. ZSFDet synthesizes discriminative and structured features for fine-grained food classes through:

1) Knowledge-Enhanced Feature Synthesizer (KEFS): KEFS adopts a multi-source graph fusion module with a knowledge encoder and attention modules to learn the knowledge representation from multi-source graph embeddings. The knowledge representation is fused with semantic content and then decoded into structure-aware synthesized features by a fusion decoder, which can exploit the interaction between semantic content from word vectors and attributes and knowledge representation.

2) Rich Semantic Representation: Only using word vectors cannot provide rich enough semantic information for fine-grained ZSFD. Unlike existing methods that rely on only word embeddings or semantic attributes, we use both food attributes and word embeddings for handling ZSFD. We explore different fusion strategies to explore the effective interaction of attribute embeddings and word embeddings to support further robust feature synthesizing.

3) Region Feature Diffusion Model (RFDM): To make sure the features we create are diversified and realistic, we use a novel 1-D diffusion model as the core generator in KEFS. Based on the probabilistic modeling of the denoising diffusion process, our proposed RFDM can enhance the training of zero-shot detectors and improve adaptability and accuracy in real-world scenarios via robust region feature synthesis.

To evaluate the effectiveness of ZSFD, we first adapt the existing food detection dataset UECFOOD-256 into a ZSFD dataset. However, most images in UECFOOD-256 contain only a single object, which is not appropriate to challenge detection algorithms. Moreover,  UECFOOD-256 lacks multi-object real-world images with fine-grained attribute annotation. As attributes are required as side information for the fine-grained ZSFD task \cite{badirli2021fine}, we present a real-world dataset Food Objects With Attributes (FOWA), which contains 20,603 images collected in 10 real-world restaurant scenarios and 95,322 bounding box annotations. It is noted that FOWA has rich attributes annotated using the food knowledge graph (e.g., FoodKG \cite{haussmann2019foodkg}). The final established FOWA has 20,603 images and 228 classes with 235 attributes for each class. Compared with previous datasets, FOWA has a much richer bounding box and attribute annotation, which is essential for ZSFD.

To summarize, major contributions in this paper include the following three aspects: 
\begin{itemize}
    \item  We propose a novel ZSFD framework ZSFDet that exploits the semantic interaction between attributes, word vectors and multi-source graphs, and learns effective zero-shot detectors via synthesizing knowledge-enhanced features that are inter-class separable.
    \item  We benchmark the real-world ZSFD task and provide fine-grained attribute annotations in our established dataset FOWA. With challenging fine-grained categories and attribute annotations, our FOWA can also facilitate future research on ZSD.
    \item  Extensive experiments show that our proposed ZSFDet achieves superior performance on the FOWA and UECFOOD-256. Moreover, ZSFDet also shows effectiveness on popular ZSD datasets PASCAL VOC and MS COCO, indicating good generalization ability.
\end{itemize}

%% file: 2_related_work.tex
\section{Related Work}
\label{sec:related}

\subsection{Food Detection}
With the development of food computing \cite{survey}, various research topics such as food recognition \cite{min2023large, bossard2014food}, food detection \cite{aguilar2018grab} and multimodal food learning~\cite{wang2019learning, marin2019recipe1m} are bringing new challenges and perspectives to computer vision under real-world settings. As a fundamental technique in food computing, food detection locates and recognizes food objects, and further provides support for health-relevant applications~\cite{wang2022review}. Advances in general object detection promote early explorations in food detection~\cite{rachakonda2020ilog}. Recently, researchers have been working on domain-adaptive methods to tackle particular issues in food images (e.g., fine-grained problems between food classes) and overcome limitations of data and annotations \cite{Ege2018Multi, shimoda2019webly}. Although promising results have been achieved via domain-specific food detection methods, it is still frustrating that high-performance food detection models can barely handle real-world tasks. For example, the Food-specific DETR in Table~\ref{tab:bench} is trained following the strategies in \cite{min2023large}. Though it has achieved nearly 95\% AP50 on two food detection datasets, UNIMIB2016~\cite{ciocca2016food} and Oktoberfest\cite{ziller2019oktoberfest}, the overall accuracy of it was still less than 20\% when we deployed it in specific restaurant scenarios. A major reason is that meals of new classes update constantly in real-world scenarios, and food detection models trained with fixed seen classes can not deal with unseen food objects. To address this issue, we first build the Zero-Shot Food Detection (ZSFD) framework for detecting both seen and unseen food objects. Considering there are no available datasets for ZSFD with attribute annotations, we present the FOWA dataset with 235 attributes to evaluate ZSFD performance.

\begin{table}[!t]
  \centering
  \caption{The food detection results on UNIMIB2016 and Oktoberfest datasets (\%).}
    \setlength{\tabcolsep}{3.7mm}{
    \begin{tabular}{lcccc}
    \toprule
    \multirow{2}[4]{*}{Model} & \multicolumn{2}{c}{UNIMIB2016} & \multicolumn{2}{c}{Oktoberfest} \\
\cmidrule{2-5} & mAP & AP50 & mAP & AP50 \\
    \midrule
		EfficientDet-D3\cite{tan2020efficientdet} & 83.6 & 90.4  & 66.3 & 91.3 \\
		Cascade R-CNN\cite{cai2018cascade} & 72.4 & 86.5  & 59.7 & 86.1 \\
		DETR\cite{carion2020end} & 72.2 & 84.2 & 58.1 & 84.0 \\
		Deformable DETR\cite{zhu2020deformable}  & 83.9 & 90.7 & 63.4 & 93.6  \\
        \textbf{Food-specific DETR} & \textbf{86.6} & \textbf{94.1} & \textbf{68.2} & \textbf{94.5} \\
    \bottomrule
    \end{tabular}}
  \label{tab:bench}%
\end{table}%

\subsection{Zero-Shot Learning}
Zero-Shot Learning (ZSL) has become an emerging research field that recognizes objects whose instances are not seen during training based on word vectors or semantic attributes as side information \cite{lampert2013attribute, norouzi2014zero}.
Early ZSL exploration is dominated by mapping-based methods that embed features into the same space and search the nearest neighbor in the embedding space for input samples. Mapping-based ZSL methods can be broadly divided into three types according to the embedding space: mapping from visual space to semantic space \cite{socher2013zero, lei2015predicting}, mapping from semantic space to visual space \cite{zhang2017learning, kodirov2017semantic}, or mapping the visual features and semantic vectors into a common latent space \cite{akata2016multi}. However, since models of mapping-based methods are only trained with seen features, these models can have a severe bias against seen classes in Generalized Zero-Shot Learning (GZSL), where both seen and unseen classes appear at test time. To address this, generation-based methods provide a new strategy that synthesizes features for unseen classes and transforms the ZSL problem into traditional supervised learning. Although the generation-based methods achieve better performance in realistic GZSL tasks, they still suffer from the hypersensitivity towards the correlation between side information and visual attributes \cite{schonfeld2019generalized}. In this paper, our framework is built based on generation-based methods since the real-world ZSFD scenarios require stronger GZSL ability. We extract fine-grained attributes from the food knowledge graph to tackle the hypersensitivity problem caused by the dependency on rich semantic information.

\begin{figure*}[t!]
	\centering
	\includegraphics[width=18cm]{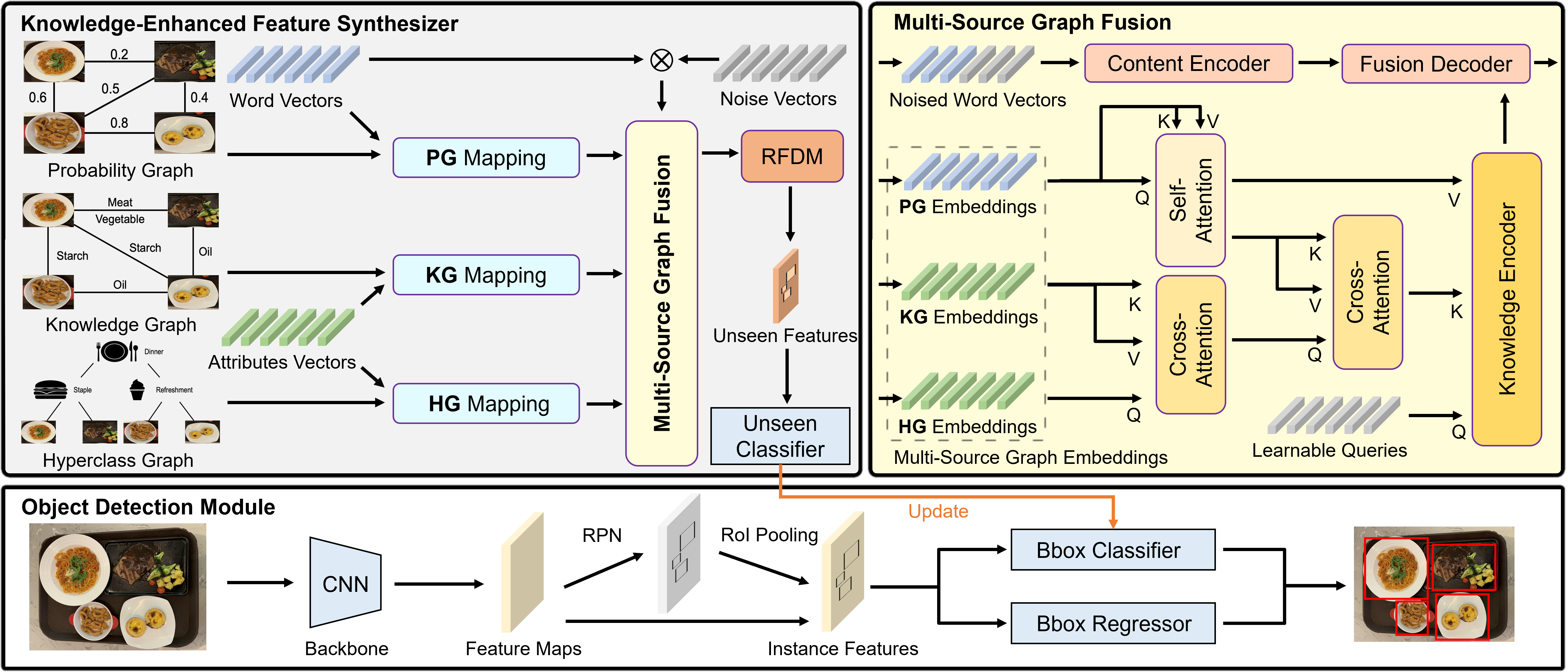}
	\caption{Overview of our ZSFDet. The ZSFDet framework contains the Knowledge-Enhanced Feature Synthesizer (KEFS) and an object detection module. The object detector is first trained on the labeled seen data, and then the unseen classifier is trained on synthesized unseen features. $\otimes$ denotes the concatenation operation, “PG” denotes the probability graph, “KG” denotes the knowledge graph, “HG” denotes the hyperclass graph, and “RFDM” denotes the Region Feature Diffusion Model.}
	% \vspace{-0.2cm}
\label{fig:approach}
\end{figure*}

\subsection{Zero-shot Object Detection}
Based on the theory of ZSL, Zero-Shot Detection (ZSD) detects objects whose instances are unseen during training\cite{xian2018zero, dai2023synthetic}. Earlier ZSD methods use mapping functions to align visual and semantic features~\cite{rahman2018zero, li2019zero, zheng2020background}, but they suffer from bias against seen classes in Generalized ZSD (GZSD), where both seen and unseen classes appear at test time. To address this, generation-based methods use generative models (e.g., Variational Autoencoder (VAE) and Generative Adversarial Networks (GAN)) to synthesize visual features for unseen classes and transform the ZSD problem into traditional supervised learning~\cite{zhu2020don,zhao2020gtnet,huang2022robust}. For example, Zhu \textit{et al.}~\cite{zhu2020don} propose an unseen feature generation framework based on VAE. Despite better GZSD performance, generation-based ZSD methods face the problem of synthesizing features that are less dispersed than real features, which limits further improvement. The latest models relieve this problem by designing loss functions that ensure the diversity of generated samples. For example, Hayat \textit{et al.}~\cite{hayat2020synthesizing} synthesize unseen features via GAN with the diversity regulation. Huang \textit{et al.}~\cite{huang2022robust} design a robust region feature synthesizer for generating diverse features. However, current ZSD methods often struggle to perform well in specific fine-grained scenarios, such as in food detection across broad categories~\cite{badirli2021fine}. The subtle differences and vast variety in food instances make it challenging to create accurate and diverse synthetic features for ZSFD. Therefore, current research aims to bridge this gap by modeling complex domain knowledge using fine-grained semantic or visual feature correlation. For example, ingredient correlation from food knowledge can be exploited by multi-source graph fusion for synthesizing discriminative features to distinguish food classes with complex attributes. It is worth noting that Li \textit{et al.}~\cite{li2023zero} also use graph neural networks for modeling the correlation between object regions in images. Different from it, the graph fusion module in our proposed ZSFDet works as style supervision in feature generation that ensures the distribution of generated features is close to the real distribution via graph regulation.

%% file: 3_method_def.tex
\section{Method}
\label{sec:method}

\subsection{Problem Definition}
\label{subsec:problem}

We formally define Zero-Shot Food Detection (ZSFD). Assuming a training set $\mathcal{X}_s$ that totally includes $M_s$ food images belonging to $C_s$ seen classes is provided. $\mathcal{Y}_s = \{1, ..., C_s\}$ and $\mathcal{Y}_u = \{C_s+1, ..., C\}$ are label sets of seen classes and unseen classes, where $\mathcal{Y}_s \cap \mathcal{Y}_u = \emptyset$, $C = C_s + C_u$ is the number of all classes, and $C_u$ is the number of unseen classes. $\mathcal{Y} = \mathcal{Y}_s \cup \mathcal{Y}_u$ denotes the total label set. Let semantic vector set be $\mathcal{V} = \mathcal{V}_s \cup \mathcal{V}_u$, where $\mathcal{V}_s$ and $\mathcal{V}_u$ are semantic vector sets of seen and unseen classes. For each $y\in\mathcal{Y}$, the semantic vector $\bm{v}_y \in \mathcal{V}$ can be both attribute vector or word vector from language models (e.g., BERT \cite{DevlinCLT19}). During the inference, a test set $\mathcal{X}_t$ that contains both $C_s$ seen classes and $C_u$ unseen classes is given. The aim of ZSFD is to learn a detector on $\mathcal{X}_s$ with semantic vectors and detect unseen objects on $\mathcal{X}_t$. ZSFD also evaluates methods on a ZSD unseen set $\mathcal{X}_u\subset \mathcal{X}_t$ that only contains $C_u$ unseen classes.

%% file: 4_method_main.tex
\subsection{The Framework Overview}
\label{subsec:overview}

As illustrated in Figure \ref{fig:approach}, our framework ZSFDet contains the proposed Knowledge-Enhanced Feature Synthesizer (KEFS) and an object detection module. KEFS, denoted as $G(\cdot)$ is designed to understand and synthesize fine-grained food features through rich culinary knowledge, enabling our model to detect novel food objects that have no instances in training set. The detection module  $\omega_{d}$ is a two-stage object detector with a feature extractor (backbone, Region Proposal Network, and RoI pooling layer) and a detection head that contains a bounding box regressor and a classifier.

We first train the detector on seen set $\mathcal{X}_s$ with corresponding images and ground-truth annotations. Then the trained detector is used to extract region features of instances from images in $\mathcal{X}_s$. Simultaneously, word vectors and attribute vectors from the Food Knowledge Graph (FKG) are obtained corresponding to seen classes $\mathcal{Y}_s$. For class set $\mathcal{Y}$, three graphs are separately defined, including a knowledge graph that models ingredient correlation, a probability graph that models label co-occurrence probability, and a hyperclass graph that models hyperclass relationship. Within KEFS, Multi-Source Graph Fusion (MSGF) learns the semantic knowledge representations from multi-source graph embeddings, and Region Feature Diffusion Model (RFDM) generates diverse unseen features based on the semantic knowledge representations. Applying the latest diffusion model on the fusion of knowledge representations and semantic content, KEFS can learn the robust mapping from semantic space to visual space, and synthesize discriminative features that are well-separated among fine-grained classes. An unseen classifier is further trained on synthesized unseen features. Finally, parameters in the detector are updated from parameters of unseen classifiers to gain the zero-shot detection ability. The details of MSGF and RFDM in KEFS will be discussed below.
% Moreover, we implement Graph Denoising Loss (GDL) $L_G$ during training to further constrains knowledge representations to the feature distribution regularized by prior knowledge.

\subsection{Multi-Source Graph Fusion}
\label{subsec:msgf}
In the Multi-Source Graph Fusion process, we start by converting food semantic vectors (word vectors and attribute vectors corresponding to food class) into a graph that represents how different food types relate to each other. In this graph, each point (node) is a food semantic vector connected by its correlation scores with other nodes (edges). A knowledge encoder with multiple attention modules is used to fuse multi-source graph embeddings into knowledge representation. In the other branch, word vectors concatenated with noise vectors (to ensure variations in the generated features) are sent to the content encoder to obtain the semantic content. Finally, the semantic content vectors and graph representation vectors are fused and decoded into visual features by the fusion decoder. 

\noindent\textbf{Multi-Source Graphs.}
We define $\mathcal{A}=\{\bm{A}^1,\bm{A}^2,\bm{A}^3\}$ as the set of adjacency matrices for multi-source graphs, where each adjacency matrix contains edges information in graphs that model the prior class correlations between nodes. We first take each food class and analyze its ingredients and characteristics, forming knowledge adjacency matrix $\bm{A}^{1}$ that models the attribute correlation in the knowledge graph. Next, we look at how these classes are grouped in broader categories to form a hyperclass adjacency matrix $\bm{A}^{2}$ that models hyperclass relationship and a probability adjacency matrix $\bm{A}^{3}$ that models label co-occurrence probability. Employing these correlations helps our model not only recognize foods it has seen before, but also make guesses about unfamiliar foods by their 'culinary context' - much like humans make guesses about the taste of unfamiliar dishes by known similar dishes.
 
First, the attribute correlation from the knowledge graph is modeled by $\bm{A}^{1}$. On ZSFD, $\bm{A}^{1}$ represents the ingredient correlation of food classes in FKG. For example, if the $i$-th class \emph{Steak} and the $j$-th class \emph{Dried Beef} share $r$ same ingredients that belong to the same group (e.g., \emph{Beef Products}) in FKG, then entry $\bm{A}^{1}_{i,j} = r$. It means the value of edge that links the $i$-th and $j$-th nodes, which is correlation score, is assigned as $r$.

Second, the hyperclass relationship of classes is modeled by $\bm{A}^{2}$ of the hyperclass graph. Given class hierarchies in dataset, specific classes $c_{i}$ can be inserted into a tree data structure as leaves. Assuming that the class tree has $e$ levels, and the node at level $0$ is the root, which is not considered as a cursor. The cursor of two leaf classes $c_{i}$ and $c_{j}$ is defined as the same ancestor node at the highest level. Therefore, the entry $\bm{A}_{i,j}^{2}$ in $\bm{A}^{2}$ is defined as:
\begin{equation}
\bm{A}_{i,j}^{2} =\left\{
\begin{aligned}
& l,\, \text{if $c_{i}$ and $c_{j}$ have the cursor at the level $l$} \\
& 0,\, \text{if $c_{i}$ and $c_{j}$ do not have the cursor}
\end{aligned}
\right. .
\end{equation}

Finally, we construct $\bm{A}^{3}$ to model conditional probability on training data and describe the statistical correlation of labels. Let $\bm{O}_{i,j}$ be the number of occurrence times of the pair of the $i$-th and the $j$-th class labels and $\bm{T}_i$ be the occurrence times of the $i$-th label, we can obtain $\bm{A}^3_{i,j}= \bm{O}_{i,j}/\bm{T}_i$. All $\bm{A}^{1}$, $\bm{A}^{2}$ and $\bm{A}^{3}$ are normalized and quantized into logical matrices by a threshold $\tau$:

\begin{equation}
\bm{A}_{i,j}^{k} =\left\{
\begin{aligned}
& 1,\quad \text{if $\bm{A}_{i,j}^{k} \geq \tau$} \\
& 0,\quad \text{if $\bm{A}_{i,j}^{k} < \tau$}
\end{aligned}
\right. ,
\end{equation}
where $\bm{A}^{k}$ is the $k$-th prior graph in set $\mathcal{A}$. We can further utilize various semantic knowledge in multiple graphs by multi-source graph mapping.

\noindent\textbf{Multi-Source Graph Mapping.}
Multiple graph embeddings are obtained via the graph mapping $\psi^{k}(\cdot)$, which is implemented as graph convolution~\cite{KipfW17}. Let input semantic vectors $\bm{V}\in \mathbb{R}^{n\times d}$ be word vectors or attribute vectors, and the corresponding graph adjacency matrix be $\bm{A}^{k}\in \mathbb{R}^{n\times n}$, where $n$ is the number of classes and $d$ is the dimension of the semantic vector. Through two graph convolution layers, graph embeddings $\bm{E}^{k} \in \mathbb{R}^{n\times d} $ of the corresponding $k$-th prior graph in set $\mathcal{A}$ can be obtained:
\begin{equation}
\begin{aligned}
\bm{E}^{k} = \psi^{k}(\bm{V}) =  \hat{\bm{A}}^{k} ( \rho(\hat{\bm{A}}^{k} \bm{V} \bm{W}_1^{k}) ) \bm{W}_2^{k}
\end{aligned}
,
\end{equation}
where $\rho(\cdot)$ denotes LeakyReLU \cite{maas2013rectifier}, 
$\hat{\bm{A}_k} = \bm{D}_k^{-\frac{1}{2}} \bm{A}_k \bm{D}_k^{-\frac{1}{2}}$ is  Laplacian normalized version of the correlation matrix, and $\bm{D}_k$ is the diagonal node degree matrix of $\bm{A}_k$. $\bm{W}_1^{k}\in \mathbb{R}^{d\times d'}$ and $\bm{W}_2^{k}\in \mathbb{R}^{d'\times d}$ are transformation matrices, and $d'$ is the latent dimension.

\begin{figure}[t]
	\centering
  \includegraphics[width=1.01\columnwidth]{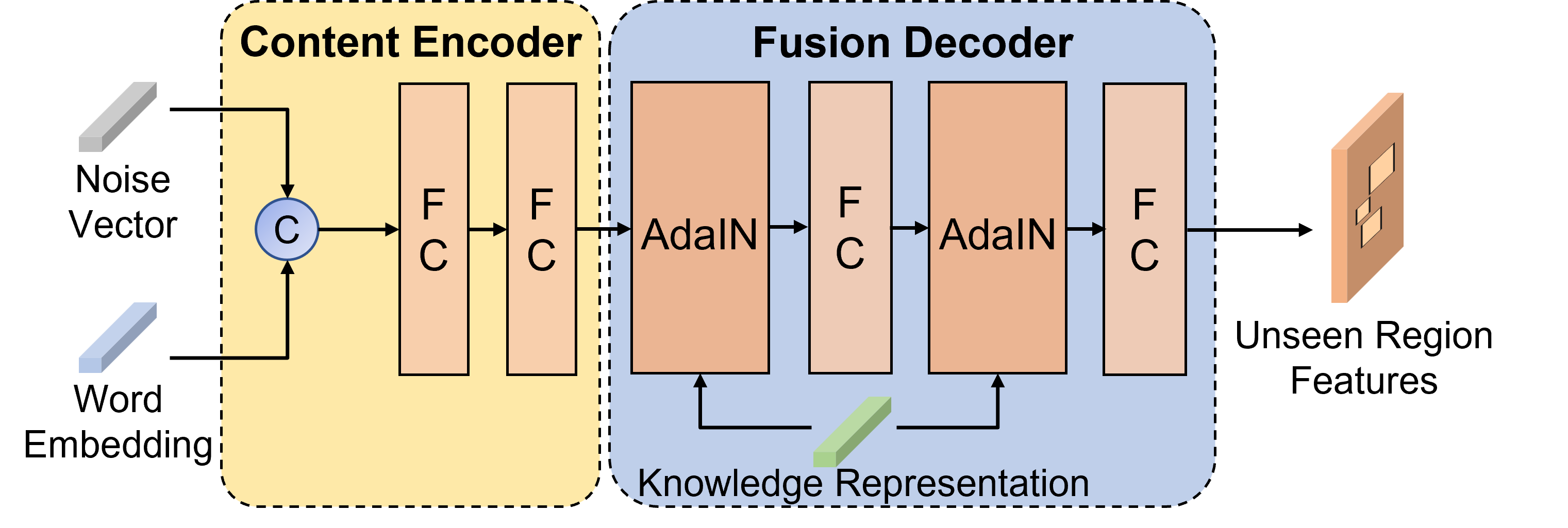}
  \caption{Illustration of details in the content decoder and fusion decoder modules.}
  \label{fig:encdec} 
  % \vspace{-0.3cm}
\end{figure}

\noindent\textbf{Knowledge Encoder.}
The  knowledge encoder $\varphi(\cdot)$ encodes the fused word and attribute graph embeddings $\bm{E}_{f}  \in \mathbb{R}^{n\times z}$ and word graph embeddings $\bm{E}_{w2v} \in \mathbb{R}^{n\times z}$ from word vectors $\bm{V}_{w2v}$ into knowledge representation vectors $\bm{S} \in \mathbb{R}^{n \times d}$:
\begin{equation}
\begin{aligned}
\bm{S} = \varphi(\bm{Q}, \bm{E}_{f}, \bm{E}_{w2v}) = 
% \text{Softmax}\left(\frac{\bm{Q}\bm{W}_i^Q (\bm{E}_{f}\bm{W}_i^K)^T}{\sqrt{d_q/h}}\right)(\bm{E}_{w2v}\bm{W}_i^V),
\text{MHA}(\bm{Q}\bm{W}_Q, \bm{E}_{f}\bm{W}_K, \bm{E}_{w2v}\bm{W}_V),
\end{aligned}
\end{equation}
where $z$ is the dimension of embedded features and $\bm{Q} \in \mathbb{R}^{n \times z}$ is a set of learnable queries that are randomly initialized by Gaussian distribution $\mathcal{N}(\mathbf{0},\mathbf{I})$. The key $\bm{E}_{f}$ is obtained by fusing the word and attribute graph embeddings via cross-attention modules, and the value $\bm{E}_{w2v}$ is obtained by encoding probability graph embeddings $\bm{E}_{w2v}^{3} \in \mathbb{R}^{n \times d}$ via self-attention modules. $\bm{W}_Q  \in \mathbb{R}^{z\times d}$, $\bm{W}_K\in \mathbb{R}^{z\times d}$ and $\bm{W}_V\in \mathbb{R}^{z\times d}$ are corresponding weight matrices for query, key and value. MHA$(\cdot)$ denotes the multi-head attention mechanism.

\noindent\textbf{Content Encoder and Fusion Decoder.}
Noise vectors $\bm{Z} \in \mathbb{R}^{n \times d}$ are sampled from the Gaussian distribution $\mathcal{N}(\mathbf{0},\mathbf{I})$. As illustrated in Figure~\ref{fig:encdec}, the content encoder consists of two linear layers. It takes the concatenation of input word vectors $\bm{V}_{w2v}$ and noise vectors $\bm{Z}$ as input and output content vectors $\bm{N} \in \mathbb{R}^{n \times e}$, where $e$ is the new embedding dimension:
\begin{align}
\bm{N} = \mathrm{FC2}(\mathrm{FC1}(\bm{V}_{w2v} \otimes \bm{Z})),
\end{align}
where $\otimes$ denotes the concatenation operation, and $\mathrm{FC1}$ and $\mathrm{FC2}$ denote linear operations followed by LeakyReLU.

The fusion decoder leverages two Adaptive Instance Normalization (AdaIN) \cite{huang2017arbitrary} blocks with two linear transformations to normalize the content with the distribution of knowledge representation:
\begin{align}
\text{AdaIN}(\bm{N}, \bm{S}) =  \sigma(\bm{S}) \left( \frac{\bm{N}-\mu(\bm{N})}{\sigma(\bm{N})} \right) + \mu(\bm{S}),
\end{align}
where $\mu(\cdot)$ and $\sigma(\cdot)$ are the mean and standard deviation. Finally, the last fully connected layer projects fusion embeddings into the synthesized instance feature $\bm{H} \in \mathbb{R}^{n\times a}$, where $a$ is the dimension of visual instance features.

\noindent\textbf{Graph Denoising Loss.}
To ensure that the learned knowledge representations are constrained to approach the distribution and regularized by prior graphs, we propose the graph denoising loss $\mathcal{L}_G$ based on the graphs defined in MSGF, which can ensure the diversity of synthesized features when conditioned on the knowledge representation $\bm{S}$:

\begin{equation}
\mathcal{L}_{G} =\mathbb{E}\left[- \frac{1}{C} \sum_{k=1}^{3}\sum_{i=1}^{C} {y}_i log(\hat{\bm{s}}_i) - \alpha \hat{\bm{s}}_i log(\phi(\bm{b}^{k}_i )) \right],
\end{equation}
where ${y}_i$ is the $i$-th class label, $\phi(\cdot)$ is the sigmoid function, $\hat{\bm{s}}_i = \phi(\bm{s}_i)$, $\bm{s}_i \in \mathbb{R}^{n}$ is the $i$-th row vector in matrix $\bm{S}$, $\bm{b}^{k}_i \in \mathbb{R}^{n}$ is the $i$-th row vector in matrix $\bm{A}^{k}\bm{S}$, and $\alpha$ is the trade-off factor that adjusts the contributions of two terms.

\begin{figure}[t]
	\centering
	\includegraphics[trim=0 0.1cm 0 0, width=8.2cm]{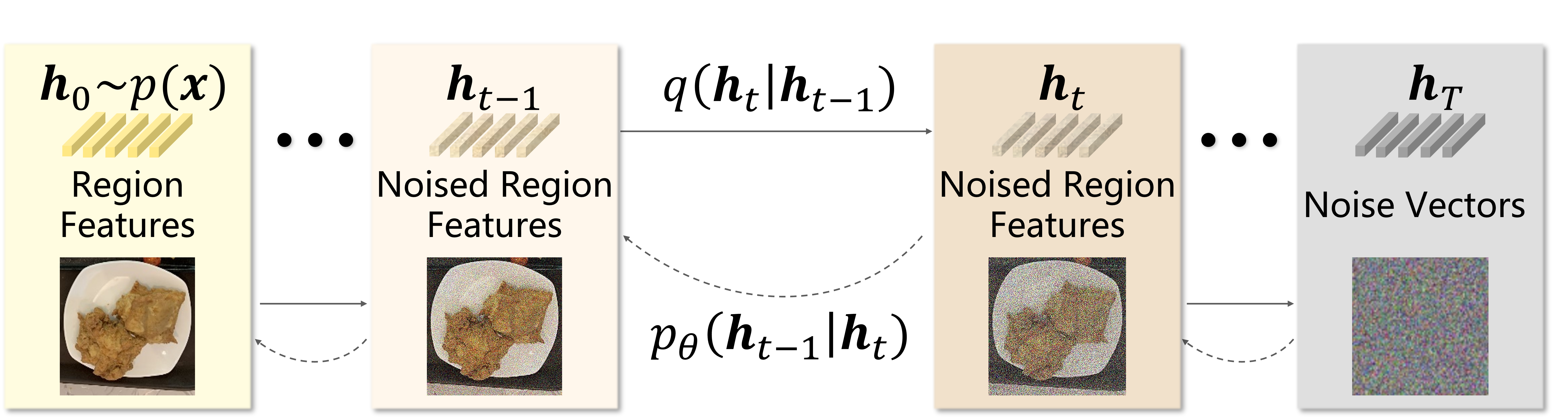}
	\caption{The visual illustration of the diffusion process in RFDM. The forward process $q(\bm{h}_t|\bm{h}_{t-1})$ continually add Gaussian noise to $\bm{h}_{t-1}$ (from right to left), the reverse process $p_\theta(\bm{h}_{t-1}|\bm{h}_t)$ aims to denoise the noised feature vector $\bm{h}_{t}$.}
	\label{fig:dif}
	% \vspace{-0.5cm}
\end{figure}

\begin{algorithm}[!t]
\caption{Training procedures of our ZSFDet}
\begin{algorithmic}[1] 
\REQUIRE $\mathcal{X}_s$, $\mathcal{X}_u$, $\mathcal{Y}_s$, $\mathcal{Y}_u$, $\mathcal{V}_{w2v}$, $\mathcal{V}_{att}$, $\mathcal{A}$ 
\ENSURE Zero-shot detector with parameters $\omega_d$ 
\STATE $\omega_d$ ← Train detector on $\mathcal{X}_s$ with annotations 
\STATE $\mathcal{H}_s$ ← Extract region features from $\mathcal{X}_s$ via $\omega_d$ 
\STATE $\bm{G}$ ← Initialize KEFS with $\mathcal{A}$ \STATE $\bm{G}$ ← Train $\bm{G}$ on $\mathcal{H}_s$, $\mathcal{V}_{w2v}$, $\mathcal{V}_{att}$, and $\mathcal{Y}_s$ by optimi-\\zing loss in Eq. \ref{loss} 
\STATE $\mathcal{H}_u$ ← Synthesize region features of unseen class-\\es using the trained $\bm{G}$, $\mathcal{V}_{w2v}$, and $\mathcal{V}_{att}$ 
\STATE $\omega_{uc}$ ← Train unseen classifier $\omega_{uc}$ using $\mathcal{H}_u$ and $\mathcal{Y}_u$
\STATE $\omega_d$ ← Update parameters in $\omega_d$ with $\omega_{uc}$ 
\RETURN $\omega_d$ 
\end{algorithmic}
\label{alg}
\end{algorithm}

\begin{figure*}[htbp]
	\begin{center}
		\includegraphics[width=0.98\textwidth]{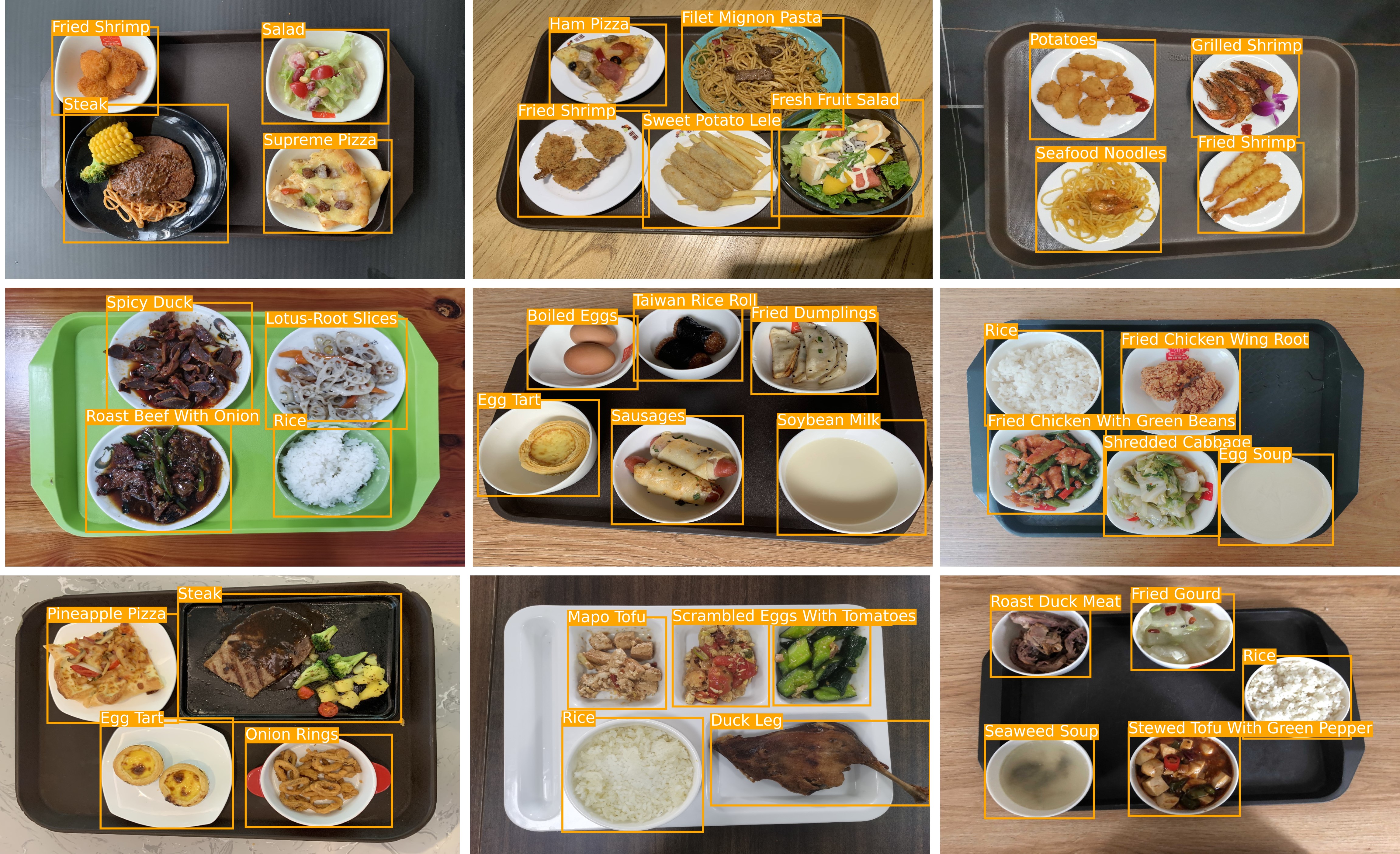}
		\caption{Visualization examples of images and bounding boxes annotations in FOWA.}
		\label{fig:morex}
	\end{center}
\end{figure*}

\begin{figure*}[htbp]
	\begin{center}
		\includegraphics[width=1.01\textwidth]{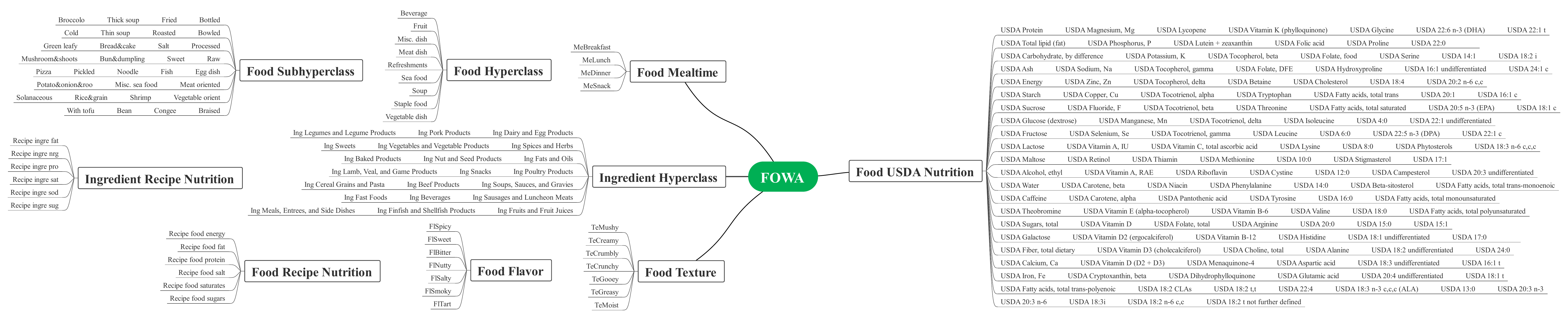}
		\caption{Illustration of attributes in FOWA. Attributes in FOWA can be categorized into 9 hyperclasses.}
		\label{fig:att}
	\end{center}
\end{figure*}

\subsection{Region Feature Diffusion Model}
Our ZSFDet uses a new generator named Regional Feature Diffusion Model (RFDM). The RFDM can generate diverse one-dimensional (1D) region feature vectors, closely mimicking variability in real-world culinary environments and enriching the diversity of synthesized food features. As Figure~\ref{fig:dif} illustrates, RFDM is based on the concept of modeling data distribution as a diffusion process, beginning from a basic prior distribution and progressively incorporating noise. The method learns to generate samples through a sequence of noise removal steps, reversing the diffusion process and retrieving the region feature.

Consider $\bm{h} \in \mathbb{R}^{d}$ as a 1D region feature vector. We propose $\bm{h}_T$ is generated through a diffusion process starting from the sample $\bm{h}_0 \sim p_0(\bm{h})$, where $p_0$ is the data distribution, and at each timestep $t = 1, …, T$, Gaussian noise is added following a Markovian process. The noise level at each timestep is regulated by a scalar $\gamma_t \in (0, 1)$. The forward diffusion process $q(\bm{h}_t|\bm{h}_{t-1})$ for each timestep is described as:

\begin{equation}
\bm{h}_t = \sqrt{1 - \gamma_t} \bm{h}_{t-1} + \sqrt{\gamma_t} \bm{z}_t ,
\end{equation}
where $\bm{z}_t  \in \mathbb{R}^{d}$ is sampled from $\mathcal{N}(\mathbf{0},\mathbf{I})$ and $\bm{h}_0 = \bm{h}$. The RFDM seeks to reverse this process, represented by:

\begin{equation} 
p_{\theta}(\bm{h}_{t-1}|\bm{h}_t) = \mathcal{N}(\bm{h}_{t-1}|\mu_{\theta}(\bm{h}_t, t), \Sigma_{\theta}(\bm{h}_t, t)\mathbf{I}),
\end{equation} 
where $\bm{h}_{t-1}$ and $\bm{h}_t$ represent the feature vector at timestep $t-1$ and $t$, respectively. $\mathcal{N}(\cdot)$ denotes Gaussian distribution, $\Sigma_{\theta}(\bm{h}_t, t)$ is a fixed covariance and $\mu_{\theta}(\bm{h}_t, t)$ is a predicted mean by RFDM. Specifically, RFDM uses the knowledge representations from MSGF as conditions to predict the noise for recovering the region feature based the predicted mean:

% \begin{equation}
%     \bm{\mu}_{\theta}(\bm{h}_t, t) = \frac { \sqrt { \beta_t } (1 - \bar{\beta}_{t-1}) \bm{h}_t + \sqrt{ \bar{\beta}_{t-1} } (1 - \beta_t) \bm{z}(\bm{h}_t, t)}   {1 - \bar{\beta}_t} ,
% \end{equation}
\begin{equation}
    \bm{\mu}_{\theta}(\bm{h}_t, t) = \frac {1} {\sqrt{\beta_t}} ( \bm{h}_t - \frac{1-\beta}{\sqrt{ 1 - \bar{\beta}_{t} }} \bm{z}_\theta (\bm{h}_t, t) ),
\end{equation}
where $\beta_t = 1 - \gamma_t$, $\bar{\beta}_t = \prod^t_{i=1} \beta_i$ is the accumulated noise scalars, and $\bm{z}_\theta(\bm{h}_t, t)$ is the predicted noise parameterized by RFDM. Thus we can map $\bm{h}_t$ to $\bm{h}_0$ by applying a series of denoising functions $\bm{G}_t$:
\begin{equation}
\bm{h}_{t-1} = \bm{G}_t(\bm{h}_t, t, \bm{z}_\theta(\bm{h}_t, t); \theta) ,
\end{equation}
where $\theta$ are the parameters of the MSGF module in RFDM. The denoising functions $\bm{G}_t$ are implemented by MSGF modules that share the same architecture but have different parameters for each timestep. The RFDM is trained by minimizing the mean squared error between the real noise $\bm{z}_t$ and the predicted noise $\bm{z}_\theta(\bm{h}_t, t)$ for all timesteps:

\begin{equation}
\begin{aligned}
\mathcal{L}_R &= \mathbb{E}_{\bm{h},\bm{z}_t}[\sum_{t=1}^T ||\bm{z}_{t} - \bm{z}_\theta(\bm{h}_t, t)||^2] \\
&=  \mathbb{E}_{\bm{h},\bm{z}}[\sum_{t=1}^T ||\bm{z}_{t} -  \bm{G}_t(\bm{h}_t, t, \bm{z}_\theta(\bm{h}_t, t); \theta)||^2] .
\end{aligned}
\end{equation}

\begin{figure*}[htbp]
	\begin{center}
		\includegraphics[width=1\textwidth]{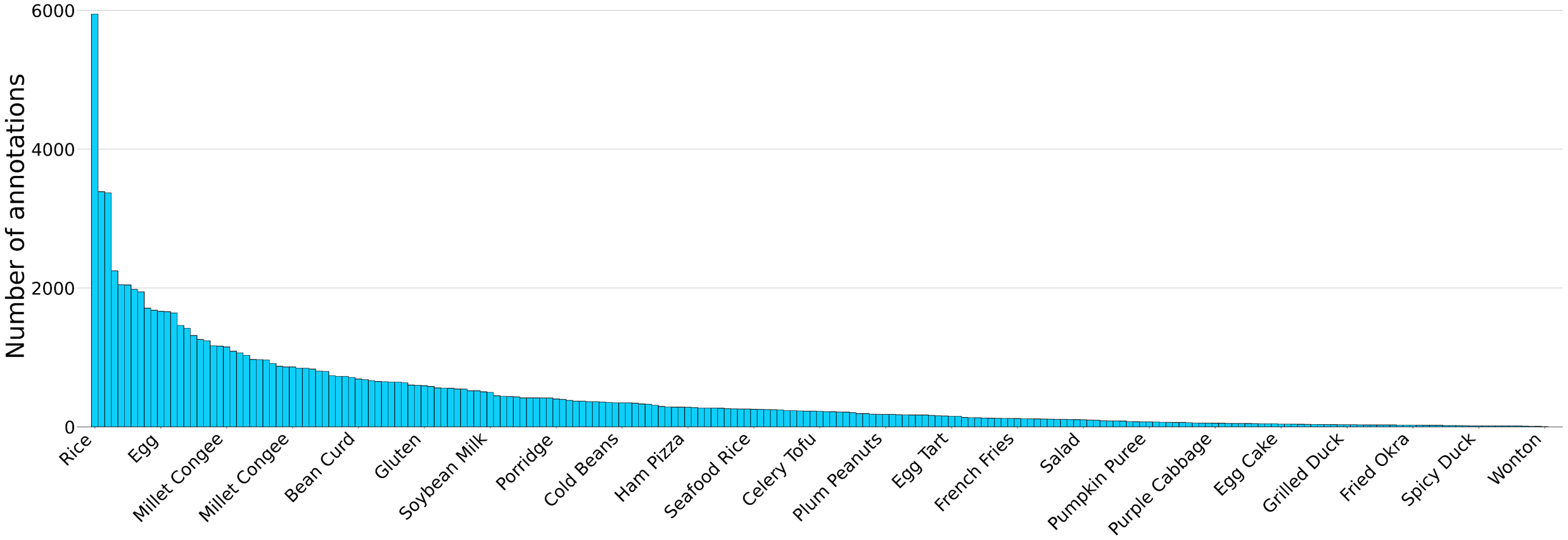}
		\caption{Sorted numbers of bounding box annotations distribution in FOWA. Sampled class names are on the x-axis and the numbers of annotations are on the y-axis.}
		\label{fig:stat}
	\end{center}
 % \vspace{-0.3cm}
\end{figure*}

\subsection{Training Objective of KEFS}
\label{sec:loss}

Given the seen instance feature collection $\mathcal{H}_s$ with word vector set $\mathcal{V}_{w2v}$ and attribute vector set $\mathcal{V}_{att}$ from $\mathcal{X}_s$, our goal is to learn a synthesizer $\bm{G}$: $(\mathcal{V}_{w2v} \times \mathcal{V}_{att} \times \mathcal{Z}) \mapsto \mathcal{H}$. The synthesizer $\bm{G}$ takes a word vector $\bm{v}_{vec} \in \mathcal{V}_{vec}$, an attribute vector $\bm{v}_{att} \in \mathcal{V}_{att}$ and a noise vector $\bm{z} \in \mathcal{Z}$ from the Gaussian distribution as input, and outputs the synthesized instance features $\tilde{\bm{h}} \in \mathcal{H}$. Specifically, the total training objective in KEFS comprises three parts: the conditional Wasserstein generative loss $\mathcal{L}_{W}$~\cite{arjovsky2017wasserstein} that is used to learn the knowledge representation as the condition of RFDM, the reconstruction loss $\mathcal{L}_R$ of RFDM and the Graph Denoising Loss $\mathcal{L}_G$ :

\begin{align}
	\mathcal{L}_{Total} =\mathop{\min}_{G} \mathop{\max}_{D} \ & \mathcal{L}_{W} + \lambda_1 \mathcal{L}_{R} + \lambda_2 \mathcal{L}_{G},
\label{loss}
\end{align}
where $\lambda_{1}$ and $\lambda_2$ are weights of losses. Training procedures of our method are summarized in Algorithm \ref{alg}.

%% file: 5_dataset.tex
\section{The FOWA Dataset}
\label{sec:avai}

% Table generated by Excel2LaTeX from sheet 'datasets'

% \begin{figure}[t!]
% 	\centering
% 	\includegraphics[width=0.45\textwidth]{data.pdf}
% 	\caption{The example of FOWA. Food images with attributes of categories are presented .}
% 	\label{fig:label_statistic}
% \end{figure}

In the FOWA dataset, images are collected in 10 restaurant scenarios and attributes are extracted from the food knowledge graph FoodKG \cite{haussmann2019foodkg}. Attributes from FoodKG include various ingredients and nutritional data of food entities. Since FoodKG contains massive food entities, it is difficult for humans to link classes in real food images to FoodKG. Thus, we first use the bi-directional matching \cite{zhou2022self} to efficiently match the closest food entities in FoodKG to classes in our dataset and form a similar list. We further use BERT to extract semantic vectors for each class and use the cosine distance to compute the similarity score between semantic vectors of entities from FoodKG and our dataset. Moreover, matched pairs with the highest similarity are filtered by a threshold $\epsilon=0.7$. Finally, 228 classes are successfully linked to entities in FoodKG and double-checked manually. As a result, the FOWA has 20,603 images and 228 classes, and each class is annotated with 235 attributes. FOWA also has three levels of classes, which include 9 hyperclasses (e.g., meat and vegetable), 21 subclasses (e.g., poultry and beef), and 228 specific classes. Since the Large Language Models (LLMs) were not yet available when the project was completed, most of the semantic attributes were manually annotated. Nowadays, leveraging advanced LLMs such as GPT-4~\cite{openai2023gpt} and GPT-4V~\cite{yang2023dawn}, we can automatically generate rich semantic attributes for hundreds of food classes directly. This technological advancement significantly reduces the laborious effort required in annotation, making the process of expanding our dataset more efficient to more comprehensively evaluate related methods.

As illustrated in Figure~\ref{fig:morex} and Figure~\ref{fig:att}, FOWA stands out with its rich attribute annotations covering a wide range from taste and texture to nutritional content, making it particularly suited for the complex tasks of Zero-Shot Food Detection (ZSFD). 235 attributes in FOWA can be categorized into 9 hyperclasses: \emph{Food Mealtime}, \emph{Food Flavor}, \emph{Food Texture}, \emph{Food Hyperclass}, \emph{Food Subhyperclass}, \emph{Ingredient Hyperclass}, \emph{Food USDA Nutrition}, \emph{Food Recipe Nutrition}, and \emph{Ingredient Recipe Nutrition}. The dataset's long-tail distribution of bounding box annotations, as presented in Figure \ref{fig:stat}, presents a realistic and challenging scenario for models to tackle. We compare the proposed FOWA to existing food datasets like UECFOOD-256 in Table \ref{tab:dataset_comparison}, where we can see that FOWA provides a significantly larger set of bounding box annotations (95,322 compared to 28,429). Unlike common object detection datasets such as MS COCO, FOWA specializes in fine-grained attribute annotations crucial for zero-shot learning, offering a more challenging and relevant benchmark for ZSFD approaches. In terms of open-source, we are committed to making FOWA widely available to the research community.
% The rich attributes provide the basis for ZSFD. To better illustrate the properties of the established FOWA, an overall statistic for all categories is presented in Figure \ref{fig:stat}. The statistic shows the number of bounding box annotations that each category owns in FOWA. We can see that the number of bounding box annotations obeys the long-tail distribution, which makes FOWA more challenging.

\begin{table}[!t]
  \centering
  \caption{Statistics of the proposed FOWA and existing datasets that are widely used for ZSD.}
  	\setlength{\tabcolsep}{0.75mm}{
    \begin{tabular}{lccccccc}
    \toprule
    \multirow{2}[4]{*}{Dataset} & \multirow{2}[4]{*}{Att.} & \multicolumn{3}{c}{Classes} & \multicolumn{3}{c}{Images} \\
\cmidrule{3-5}   \cmidrule{6-8}       &       & S     & U     & Total & Training & Test  & Total \\
    \midrule
    PASCAL VOC \cite{everingham2010pascal}   & 64 & 16    & 4     & 20    & 10,728 & 10,834 & 21,562 \\
    MS COCO \cite{coco} & 0 & 65    & 15    & 80    & 82,783 & 40,504 & 123,287 \\
    % DIOR \cite{li2020object}  & 0 & 16    & 4     & 20    & 11,725 & 11,738 & 23,463 \\
    UECFOOD-256 \cite{kawano14c} & 0  & 205    & 51     & 256    & 20,452 & 5,732 & 26,184 \\
    FOWA  & 235 & 184 & 44 & 228 & 10,463 & 10,140 & 20,603 \\
    \bottomrule
    \end{tabular}}
  \label{tab:dataset_comparison}%
\end{table}%

% A comparison of FOWA against existing datasets is shown in Table \ref{tab:dataset_comparison}. Compared with UECFOOD-256 which has only 28,429 bounding boxes, FOWA owns 95,322 rich bounding box annotations. Compared with FOWA, even though available ZSD datasets such as MS COCO have large-scale images, these datasets lack fine-grained attribute annotation, which is essential for zero-shot learning. In terms of complexity, FOWA with more categories, especially unseen categories, is more suitable to challenge ZSFD approaches. We believe that FOWA can also serve as a leading benchmark with fine-grained attributes for future studies on ZSD.

%  As seen in Table~\ref{tab:dataset_comparison}, MS COCO and DIOR have no annotated attributes, where only the word embedding of the category name can be used as semantic information. The attributes in VOC are inherited by aPY \cite{farhadi2009describing}, which has only 64 elements for each attribute vector. The provided semantic information needs to be richer and fine-grained enough for ZSD. Therefore, to evaluate the performance of ZSFD methods and provide a valuable benchmark with fine-grained attribute annotation for ZSD, we present FOWA.

%% file: 6_experiments.tex
\section{Experiments}
\label{sec:experiments}

\noindent\textbf{Dataset Splittings.}
Following the setting in \cite{bansal2018zero, rahman2018polarity}, categories in FOWA are split into 184 seen classes and 44 unseen classes, and categories in UECFOOD-256 are split into 205 seen classes and 51 unseen classes, respectively. We also compare the ZSD performance of our proposed method ZSFDet with State-Of-The-Art (SOTA) ZSD methods on widely-used datasets: PASCAL VOC 2007+2012~\cite{everingham2010pascal} and MS COCO 2014 \cite{coco} using the same dataset splitting \cite{hayat2020synthesizing, huang2022robust}, which is shown in Table \ref{tab:dataset_comparison}. Note that two different splits are adopted for MS COCO: 48/17 seen/unseen split and 65/15 seen/unseen split.

\begin{table}[!t]
  \centering
  \caption{Comparison with baselines on FOWA (\%).}
    \setlength{\tabcolsep}{1.5mm}{
    \begin{tabular}{clcccc}
    \toprule
    \multirow{2}[4]{*}{Metric} & \multirow{2}[4]{*}{Method} & \multirow{2}[4]{*}{ZSD} & \multicolumn{3}{c}{GZSD} \\
\cmidrule{4-6}          &       &       & Seen  & Unseen & HM \\
    \midrule
   \multirow{8}[2]{*}{Recall@100} & ConSE~\cite{norouzi2014zero} & 39.7 & 58.0 & 38.1 & 46.4 \\
    & PL\cite{rahman2018polarity}    &   40.1    &   53.9    &  39.6     &  45.7 \\
          & BLC~\cite{zheng2020background}  & 41.2  & 55.3   & 40.5  & 46.8 \\
          & CZSD~\cite{yan2022semantics}  &   48.0    & 86.1  &  44.8 &  58.9 \\
          & SU~\cite{hayat2020synthesizing}    & 45.3  & 82.3  & 44.1  & 57.4 \\
          & RRFS~\cite{huang2022robust}  & 48.8  & 86.6  & 47.6  & 61.4 \\
          & SeeDS~\cite{zhou2023seeds} & 52.3 & 86.7 & 49.5 & 63.2 \\
          & \textbf{ZSFDet}  & \textbf{53.5}  & \textbf{87.0}  & \textbf{50.1}  & \textbf{63.6} \\
    \midrule
   \multirow{8}[2]{*}{mAP} & ConSE~\cite{norouzi2014zero} & 0.8 & 54.3 & 0.7 & 1.4 \\
    & PL~\cite{rahman2018polarity}    &    1.0   &   50.8    &   0.7    &  1.4 \\
          & BLC~\cite{zheng2020background}  & 1.1   & 51.1   & 0.9   & 1.8  \\
          & CZSD~\cite{yan2022semantics}  &  4.0     &    81.2   &   2.1  & 4.1 \\
          & SU~\cite{hayat2020synthesizing}    & 3.9   & 79.1  & 2.3   & 4.5  \\
          & RRFS~\cite{huang2022robust}  & 4.3   & 82.7  & 2.7   & 5.2  \\
          & SeeDS~\cite{zhou2023seeds} & 5.6 & 82.7 & 3.3 & 6.3 \\
          & \textbf{ZSFDet}  & \textbf{6.1}   & \textbf{82.8}  & \textbf{3.6}     & \textbf{6.9}  \\
    \bottomrule
    \end{tabular}}
  \label{tab:FOWA}%
\end{table}%

\begin{table}[!t]
  \centering
  \caption{Comparison on UECFOOD-256 (\%).}
    \setlength{\tabcolsep}{1.5mm}{
    \begin{tabular}{clcccc}
    \toprule
    \multirow{2}[4]{*}{Metric} & \multirow{2}[4]{*}{Method} & \multirow{2}[4]{*}{ZSD} & \multicolumn{3}{c}{GZSD} \\
\cmidrule{4-6}          &       &       & Seen  & Unseen & HM \\
    \midrule
   \multirow{8}[2]{*}{Recall@100} 
   & ConSE~\cite{norouzi2014zero} & 54.4 & 50.1  & 38.2 & 43.3 \\
    & PL\cite{rahman2018polarity}    &   56.5    &    53.2   &    40.4   & 46.0 \\
          & BLC~\cite{zheng2020background}  & 58.9   &  55.3  &  43.8 &  48.9 \\
          & CZSD~\cite{yan2022semantics}  &  60.7   & \textbf{57.6}  &  45.5 & 50.8 \\
   & SU~\cite{hayat2020synthesizing}    & 61.9 & 52.5  & 52.8  & 52.6 \\
          & RRFS~\cite{huang2022robust}  &  64.8 & 54.9  & 55.1  &  55.0 \\
          & SeeDS~\cite{zhou2023seeds} & 74.0 &  55.2 & 61.4 & 58.1 \\
          & \textbf{ZSFDet} & \textbf{74.4} & 57.0 & \textbf{61.8} & \textbf{59.3} \\
   \midrule
   \multirow{8}[2]{*}{mAP} 
   & ConSE~\cite{norouzi2014zero} & 11.3 & 19.7 & 9.0  & 12.4   \\
    & PL~\cite{rahman2018polarity}    &   14.5    &  18.9    &    11.6  & 14.4 \\
          & BLC~\cite{zheng2020background}  &  19.2  & 20.5  &  15.2  & 17.5  \\
      & CZSD~\cite{yan2022semantics}  &  22.0  &   \textbf{20.8}   &   16.2  & 18.2 \\
       & SU~\cite{hayat2020synthesizing}    &  22.4   &  19.3 & 20.1  & 19.7 \\
          & RRFS~\cite{huang2022robust}  &  23.6  & 20.1  &  22.9  &  21.4 \\
          &  SeeDS~\cite{zhou2023seeds} & 27.1 &  20.2 &  26.0& 22.7\\
          & \textbf{ZSFDet}  & \textbf{27.3}   & 21.9 & \textbf{26.1} & \textbf{23.8}  \\
    \bottomrule
    \end{tabular}}
  \label{tab:UEC256}%
\end{table}%

\noindent\textbf{Implementation Details.}
We adopt the Faster-RCNN \cite{fasterrcnn} with the ResNet-101 \cite{resnet} as the backbone for fair comparisons. For training the synthesizer, Adam \cite{adam2014} is used with a learning rate of 1e-5. We use $\tau=0.4$ as the threshold for logical adjacency matrices. Following the setting in \cite{detr}, both the self-attention block and knowledge encoder in our approach include 6 layers \cite{attention}. The head number of multi-head attention is 4. We set $T = 100$ for the noise sampling process of RFDM. The linear start and the linear end for noise scalars are set to $\gamma_1$ = 8.5e-4 and $\gamma_T$ = 1.2e-2. For each unseen class, we synthesize 500/500/500/250 features for FOWA/UECFOOD-256/PASCAL VOC/MS COCO to align with the setting in baseline methods \cite{hayat2020synthesizing,huang2022robust} for fair comparison. Since the ZSFDet is robust for hyper-parameters (the fluctuation range of mAP is less than 0.1\% on three datasets under grid search), we set $\lambda_1 = 0.1$, $\lambda_2 = 0.1$ and $\alpha = 0.7$. Attributes of PASCAL VOC are provided by aPY \cite{farhadi2009describing}. Word embedding vectors of class names are extracted by BERT \cite{DevlinCLT19} for FOWA and UECFOOD-256, and by FastText \cite{mikolov2018advances} for PASCAL VOC and MS COCO.
% without meticulous tuning.

\noindent\textbf{Evaluation Metrics.}
Similar to previous works \cite{bansal2018zero, huang2022robust}, we use mean Average Precision (mAP) and Recall@100 with IoU threshold 0.5 for the evaluation of FOWA, UECFOOD-256 and PASCAL VOC. For MS COCO, we report mAP and Recall@100 with IoU thresholds of 0.4, 0.5, and 0.6, respectively. We also present per-class Average Precision (AP) to compare class-wise performance. The Harmonic Mean (HM) of seen and unseen is used for performance comparison under the setting of GZSD.

\noindent\textbf{Comparison Methods.}
We compare our method with presentative classic ZSD methods and various state-of-the-art ZSD methods. These include: 
(1) \textbf{ConSE} \cite{norouzi2014zero} is a zero-shot learning method, and is reimplemented on the Faster R-CNN model for the ZSD task. Following the idea in \cite{norouzi2014zero}, the training of the detector is on the object detection training set and inference with the semantic information;
(2) \textbf{SB} \cite{bansal2018zero} is a background-aware approach that considers external annotations from object instances belonging to neither seen nor unseen, which helps to address the confusion between unseen and background;    
(3) \textbf{DSES} \cite{bansal2018zero} is a variant of SB that does not use background-aware representations but employs external data sources (e.g., large open vocabulary) for representation of background regions;    
(4) \textbf{SAN} \cite{rahman2018zero} is a Faster-RCNN-based ZSD approach that jointly models the interplay between visual and semantic domain information (e.g. super-class information);    
(5) \textbf{HRE} \cite{berkan2018zero} is a YOLO-based end-to-end ZSD approach that learns a direct mapping from region pixels to the space of class embeddings;    
(6) \textbf{PL} \cite{rahman2018polarity} is a RetinaNet-based ZSD approach that uses polarity loss for better alignment of visual features and semantic vectors;    
(7) \textbf{BLC} \cite{zheng2020background} integrates Cascade Semantic RCNN, semantic information flow and background learnable RPN into a unified ZSD framework;    
(8) \textbf{SU} \cite{hayat2020synthesizing} proposes a unified generation-based ZSD framework based on WGAN \cite{arjovsky2017wasserstein} and diversity loss;    
(9) \textbf{CZSD} \cite{yan2022semantics} incorporates two semantics-guided contrastive learning networks to learn unseen visual representations by contrasting between region-category and region-region pairs; 
(10) \textbf{RRFS} \cite{huang2022robust} proposes an intra-class semantic diverging loss and inter-class structure preserving loss to synthesize robust region features for ZSD.
(11) \textbf{TCB}~\cite{li2023zero2} is a novel two-way classification branch network for zero-shot detection that combines static and dynamic semantic vector branches.
(12) \textbf{SeeDS}~\cite{zhou2023seeds} proposes a Semantic Separable Diffusion Synthesizer to improve zero-shot food detection by enhancing region feature generation.

\begin{table}[!t]
  \centering
  \caption{Comparison of mAP on PASCAL VOC (\%).}
    \setlength{\tabcolsep}{3.8mm}{
    \begin{tabular}{lcccc}
    \toprule
    \multirow{2}[4]{*}{Model} & \multirow{2}[4]{*}{ZSD} & \multicolumn{3}{c}{GZSD} \\
\cmidrule{3-5}          &       & Seen  & Unseen & HM \\
    \midrule
    ConSE~\cite{norouzi2014zero} & 52.1  & 59.3  & 22.3  & 32.4 \\
    SAN~\cite{rahman2018zero}   & 59.1  & 48.0    & 37.0    & 41.8 \\
    HRE~\cite{berkan2018zero}   & 54.2  & 62.4  & 25.5  & 36.2 \\
    PL~\cite{rahman2018polarity}    & 62.1  & -     & -     & - \\
    BLC~\cite{zheng2020background}   & 55.2  & 58.2  & 22.9  & 32.9 \\
    CZSD~\cite{yan2022semantics} & 65.7  & \textbf{63.2}  & 46.5  & 53.8 \\
    SU~\cite{hayat2020synthesizing}    & 64.9  & -     & -     & - \\
    RRFS~\cite{huang2022robust}  & 65.5  & 47.1  & 49.1  & 48.1 \\
    TCB~\cite{li2023zero2} & 59.3 & 61.0 & 29.8 & 40.0 \\
    \textbf{ZSFDet}  & \textbf{69.2} & 48.5  & \textbf{50.8} & \textbf{49.6} \\
    \bottomrule
    \end{tabular}}
  \label{tab:voc}%
\end{table}%

\begin{table}[t]
  \centering
  \caption{Comparison of class-wise AP and mAP for different methods on unseen classes of PASCAL VOC (\%).}
    \setlength{\tabcolsep}{2.5mm}{
    \begin{tabular}{lccccc}
    \toprule
    Method & car   & dog   & sofa  & train & mAP \\
    \midrule
    SAN~\cite{rahman2018zero} & 56.2  & 85.3  & 62.6  & 26.4  & 57.6  \\
    HRE~\cite{berkan2018zero} & 55.0  & 82.0  & 55.0  & 26.0  & 54.5  \\
    PL~\cite{rahman2018polarity} & 63.7  & 87.2  & 53.2  & 44.1  & 62.1  \\
    BLC~\cite{zheng2020background} & 43.7  & 86.0  & 60.8  & 30.1  & 55.2  \\
    SU~\cite{hayat2020synthesizing} & 59.6  & 92.7  & 62.3  & 45.2  & 64.9  \\
    RRFS~\cite{huang2022robust} & 60.1  & 93.0  & 59.7 & 49.1  & 65.5  \\
    TCB~\cite{li2023zero2} & 62.9 & 72.9 & \textbf{66.2} & 35.2 & 59.3 \\
    \textbf{ZSFDet}  & \textbf{63.7} & \textbf{94.3} & 58.0  & \textbf{60.8} & \textbf{69.2} \\
    \bottomrule
    \end{tabular}}
  \label{tab:vocclass}%
\end{table}%

\begin{table}[t]
  \centering
  \caption{ZSD performance on MS COCO (\%).}
    \setlength{\tabcolsep}{0.9mm}{
    \begin{tabular}{lcrrrr}
    \toprule
    \multirow{2}[4]{*}{Model} & \multirow{2}[4]{*}{Split} & \multicolumn{3}{c}{Recall@100} & \multicolumn{1}{c}{mAP} \\
\cmidrule{3-6}          &       & \multicolumn{1}{c}{IoU=0.4} & \multicolumn{1}{c}{IoU=0.5} & \multicolumn{1}{c}{IoU=0.6} & \multicolumn{1}{c}{IoU=0.5} \\
    \midrule
    SB~\cite{bansal2018zero}    & 48/17 & \multicolumn{1}{c}{34.5} & \multicolumn{1}{c}{22.1} & \multicolumn{1}{c}{11.3} & \multicolumn{1}{c}{0.3} \\
    DSES~\cite{bansal2018zero}  & 48/17 & \multicolumn{1}{c}{40.2} & \multicolumn{1}{c}{27.2} & \multicolumn{1}{c}{13.6} & \multicolumn{1}{c}{0.5} \\
    ConSE~\cite{norouzi2014zero} & 48/17 & \multicolumn{1}{c}{28.0} & \multicolumn{1}{c}{19.6} & \multicolumn{1}{c}{8.7} & \multicolumn{1}{c}{3.2} \\
    PL~\cite{rahman2018polarity}    & 48/17 & \multicolumn{1}{c}{-} & \multicolumn{1}{c}{43.5} & \multicolumn{1}{c}{-} & \multicolumn{1}{c}{10.1} \\
    BLC~\cite{zheng2020background}   & 48/17 & \multicolumn{1}{c}{51.3} & \multicolumn{1}{c}{48.8} & \multicolumn{1}{c}{45.0} & \multicolumn{1}{c}{10.6} \\
    CZSD~\cite{yan2022semantics} & 48/17 & \multicolumn{1}{c}{56.1} & \multicolumn{1}{c}{52.4} & \multicolumn{1}{c}{47.2} & \multicolumn{1}{c}{12.5} \\
    RRFS~\cite{huang2022robust}  & 48/17 & \multicolumn{1}{c}{58.1} & \multicolumn{1}{c}{53.5} & \multicolumn{1}{c}{47.9} & \multicolumn{1}{c}{13.4} \\
     TCB~\cite{li2023zero2} & 48/17  & \multicolumn{1}{c}{55.5} &  \multicolumn{1}{c}{52.4} &  \multicolumn{1}{c}{48.1} & \multicolumn{1}{c}{11.4}\\
    \textbf{ZSFDet}  & 48/17 & \multicolumn{1}{c}{\textbf{58.6}}   & \multicolumn{1}{c}{\textbf{54.7}} &   \multicolumn{1}{c}{\textbf{48.3}}    & \multicolumn{1}{c}{\textbf{14.0}} \\
    \midrule
    ConSE~\cite{norouzi2014zero} & 65/15 & \multicolumn{1}{c}{30.4} & \multicolumn{1}{c}{23.5} & \multicolumn{1}{c}{10.1} & \multicolumn{1}{c}{3.9} \\
    PL~\cite{rahman2018polarity}    & 65/15 & \multicolumn{1}{c}{-} & \multicolumn{1}{c}{37.7} & \multicolumn{1}{c}{-} & \multicolumn{1}{c}{12.4} \\
    BLC~\cite{zheng2020background}   & 65/15 & \multicolumn{1}{c}{57.2} & \multicolumn{1}{c}{54.7} & \multicolumn{1}{c}{51.2} & \multicolumn{1}{c}{14.7} \\
    SU~\cite{hayat2020synthesizing}    & 65/15 & \multicolumn{1}{c}{54.4} & \multicolumn{1}{c}{54.0} & \multicolumn{1}{c}{47.0} & \multicolumn{1}{c}{19.0} \\
    CZSD~\cite{yan2022semantics} & 65/15 & \multicolumn{1}{c}{62.3} & \multicolumn{1}{c}{59.5} & \multicolumn{1}{c}{55.1} & \multicolumn{1}{c}{18.6} \\
    RRFS~\cite{huang2022robust}  & 65/15 & \multicolumn{1}{c}{65.3} & \multicolumn{1}{c}{62.3} & \multicolumn{1}{c}{55.9} & \multicolumn{1}{c}{19.8} \\
     TCB~\cite{li2023zero2} & 65/15  & \multicolumn{1}{c}{62.5} &  \multicolumn{1}{c}{59.9} &  \multicolumn{1}{c}{55.1}& \multicolumn{1}{c}{13.8}\\
    \textbf{ZSFDet}  & 65/15 &    \multicolumn{1}{c}{\textbf{66.5}}   &   \multicolumn{1}{c}{\textbf{64.2}}    &    \multicolumn{1}{c}{\textbf{56.7}}   & \multicolumn{1}{c}{\textbf{20.3}} \\
    \bottomrule
    \end{tabular}}
  \label{tab:cocozsd}%
\end{table}%

% Table generated by Excel2LaTeX from sheet 'COCO GZSD'
\begin{table}[!t]
  \centering
  \caption{GZSD performance on MS COCO (\%).}
    \setlength{\tabcolsep}{1.2mm}{
    \begin{tabular}{lccccccc}
    \toprule
    \multirow{2}[4]{*}{Model} & \multirow{2}[4]{*}{Split} & \multicolumn{3}{c}{Recall@100} & \multicolumn{3}{c}{mAP} \\
\cmidrule{3-8}          &       & S     & U     & HM    & S     & U     & HM \\
    \midrule
   ConSE~\cite{norouzi2014zero} & 48/17 & 43.8  & 12.3  & 19.2  & 37.2  & 1.2   & 2.3  \\
    PL~\cite{rahman2018polarity}    & 48/17 & 38.2  & 26.3  & 31.2  & 35.9  & 4.1   & 7.4  \\
    BLC~\cite{zheng2020background}   & 48/17 & 57.6  & 46.4  & 51.4  & 42.1  & 4.5   & 8.1 \\
    CZSD~\cite{yan2022semantics} & 48/17 & 65.7  & 52.4  & 58.3  & 45.1  & 6.3   & 11.1  \\
    RRFS~\cite{huang2022robust}  & 48/17 & 59.7  & 58.8  & 59.2  & 42.3  & 13.4  & 20.4  \\
     TCB~\cite{li2023zero2} & 48/17  & \textbf{71.9} &  52.4 &  \textbf{60.6}&  \textbf{47.3} & 4.9 & 8.8 \\
    \textbf{ZSFDet} & 48/17 & 60.1 & \textbf{60.7}  & 60.4 & 42.5 &     \textbf{14.3} & \textbf{21.4} \\
    \midrule
    ConSE~\cite{norouzi2014zero} & 65/15 & 41.0  & 15.6  & 22.6  & 35.8  & 3.5   & 6.4  \\
    PL~\cite{rahman2018polarity}    & 65/15 & 36.4  & 37.2  & 36.8  & 34.1  & 12.4  & 18.2  \\
    BLC~\cite{zheng2020background}   & 65/15 & 56.4  & 51.7  & 53.9  & 36.0  & 13.1  & 19.2  \\
    SU~\cite{hayat2020synthesizing}    & 65/15 & 57.7  & 53.9  & 55.7  & 36.9  & 19.0  & 25.1  \\
    CZSD~\cite{yan2022semantics} & 65/15 & 62.9  & 58.6  & 60.7  & \textbf{40.2}  & 16.5  & 23.4  \\
    RRFS~\cite{huang2022robust}  & 65/15 & 58.6  & 61.8  & 60.2  & 37.4  & 19.8  & 26.0  \\
     TCB~\cite{li2023zero2} & 65/15  & \textbf{69.3} &  59.8 &  \textbf{64.2}&  \textbf{39.9} &  13.8& 20.5\\
    \textbf{ZSFDet}  & 65/15 &   59.3   &  \textbf{63.1}  &  61.1     &  37.5   &  \textbf{20.5}  &  \textbf{26.5} \\
    \bottomrule
    \end{tabular}}
  \label{tab:cocogzsd}%
\end{table}%

\subsection{Comparisons with the State-of-the-art}
\label{sec:main_results}

\noindent\textbf{Evaluation on FOWA.} 
Experimental results on FOWA are shown in Table~\ref{tab:FOWA}, where different ZSD methods are reimplemented for comparison. Compared with the baseline method RRFS of our framework, ``ZSD'', ``Unseen'' and ``HM'' are improved by 1.8\%, 0.9\% and 1.7\% mAP, respectively. For Recall@100, ``ZSD'', ``Unseen'' and ``HM'' are improved by 4.7\%, 2.5\% and 2.2\% mAP, respectively. Compared with the latest baseline SeeDS, the ``HM'' mAP of ``GZSD'' on our ZSFDet is higher by 0.6\% on FOWA. The improvements demonstrate the proposed ZSFDet can evidently improve the ZSFD accuracy based on a robust feature synthesizer. Specifically, it utilizes the semantic knowledge from multi-source graph embeddings to help detectors achieve better ZSFD performance. We also observe that the mAP performance of all methods is low, which denotes that the larger number of classes makes ZSFD challenging, and there is more room for developing ZSFD methods. Note that the ``Seen'' performance has not been improved due to the same backbone detector for seen classes, which will be further explained in the ablation study section.

\noindent\textbf{Evaluation on UECFOOD-256.} 
We also reimplement baseline methods on UECFOOD-256 and show ZSFD results in Table~\ref{tab:UEC256}. Compared with our baseline RRFS, ``ZSD'', ``Unseen'' and ``HM'' are improved by 3.7\%, 3.2\% and 2.4\% mAP by our ZSFDet on UECFOOD-256, respectively. For Recall@100, ``ZSD'', ``Unseen'' and ``HM'' are improved by 9.6\%, 6.7\%, and 3.3\%, respectively. Compared with the latest baseline SeeDS, the ``HM'' mAP of ``GZSD'' on our ZSFDet is higher by 1.1\% on UECFOOD-256. These results further prove that the proposed ZSFDet framework with the MSGF and RFDM modules significantly improves the accuracy and recall ratio of ZSFD significantly, especially when applied to complex ZSFD scenarios in the setting of GZSD where seen and unseen food objects are presented simultaneously. We also observe that baseline methods cannot achieve as high an mAP as in ZSD when using the same data scale, demonstrating that the proposed ZSFD presents a challenging task with more room for developing future ZSFD methods.

\noindent\textbf{Evaluations on PASCAL VOC and MS COCO.}
Experimental results on PASCAL VOC are shown in Table~\ref{tab:voc}. Our ZSFDet outperforms all baselines under the ZSD setting, increasing the mAP from 65.5\% to 69.2\% compared with RRFS \cite{huang2022robust}. Furthermore, our method obtains better performance under a more challenging setting of GZSD. The ``Seen'', ``Unseen'' and ``HM'' are improved by 1.4\%. 1.7\% and 1.5\% compared with the SOTA baseline RRFS. Results show that our method achieves a more balanced performance on the seen and unseen classes for GZSD. The class-wise AP performance on PASCAL VOC is reported in Table~\ref{tab:vocclass}. We can observe that our approach achieves the best performance in 3 out of 4 classes.

We evaluate the ZSD performance on MS COCO with different IoU thresholds of 0.4, 0.5 and 0.6. As seen in Table \ref{tab:cocozsd}, our method outperforms all baseline methods. For the ``48/17'' split, our method improves the mAP and Recall@100 from 13.4\% and 53.5\% to 14.0\% and 54.7\% at IoU=0.5 compared with RRFS. For the ``65/15'' split, our ZSFDet improves the mAP and Recall@100 from 19.8\% and 62.3\% to 20.3\% and 64.2\% at IoU=0.5. As shown in Table \ref{tab:cocogzsd}, our ZSFDet also outperforms the RRFS under the GZSD setting, where ``S'' denotes performance on seen classes and ``U'' denotes performance on unseen classes. The absolute ``HM'' performance gain of our method is 1.0\% mAP and 1.2\% Recall@100 for the ``48/17'' split, and 0.5\% mAP and 0.9\% Recall@100 for the ``65/15'' split. Compared with the lastest baseline TCB, the ``HM'' of ``GZSD'' mAP of our ZSFDet is higher by 9.6\% on PASCAL VOC, by 12.6\% on the ``48/17'' split of MS COCO, and by 5.0\% on the ``65/15'' split of MS COCO, respectively. The experimental results on general ZSD datasets also demonstrate that our model exceeds existing methods in terms of both mAP and Recall@100.

\begin{table}[t]
  \centering
  \caption{Ablation studies measured by mAP (\%).}
    \setlength{\tabcolsep}{1.2mm}{
    \begin{tabular}{cccccccc}
    \toprule
    \multicolumn{1}{c}{\multirow{2}[4]{*}{Dataset}} & \multicolumn{3}{c}{Methods} & \multirow{2}[4]{*}{ZSD} & \multicolumn{3}{c}{GZSD} \\
\cmidrule{2-4}\cmidrule{6-8}          & Att. & MSGF & RFDM   &       & S     & U     & HM \\
    \midrule
    \multirow{3.5}[4]{*}{FOWA}  &  &   &       & 4.3  & 82.7  & 2.7  & 5.2 \\
    & \checkmark     &      &       & 4.5  & 82.8  & 2.8  & 5.4  \\
    & \checkmark     & \checkmark    & & 5.6 & 82.8  & 3.3  & 6.3  \\
    & \checkmark     & \checkmark     & \checkmark     & \textbf{6.1} & \textbf{82.8} & \textbf{3.6} & \textbf{6.9} \\
    \midrule
    \multirow{3.5}[4]{*}{PASCAL VOC}      &       &       &       & 65.8  & 48.4  & 49.2  & 48.4  \\
     & \checkmark     &       &       & 66.1  & 48.5  & 49.7  & 49.1  \\
     & \checkmark     & \checkmark     &       & 68.5  & 48.5  & 50.2  & 49.3  \\
     & \checkmark     & \checkmark     & \checkmark     & \textbf{69.2} & \textbf{48.5} & \textbf{50.8} & \textbf{49.6} \\
    \bottomrule
    \end{tabular}}
  \label{tab:ablation}%
%   \vspace{-0.1cm}
\end{table}%

\subsection{Ablation Study}
\label{sec:ablation}

\begin{figure}[t]
	\centering
	\includegraphics[width=8cm]{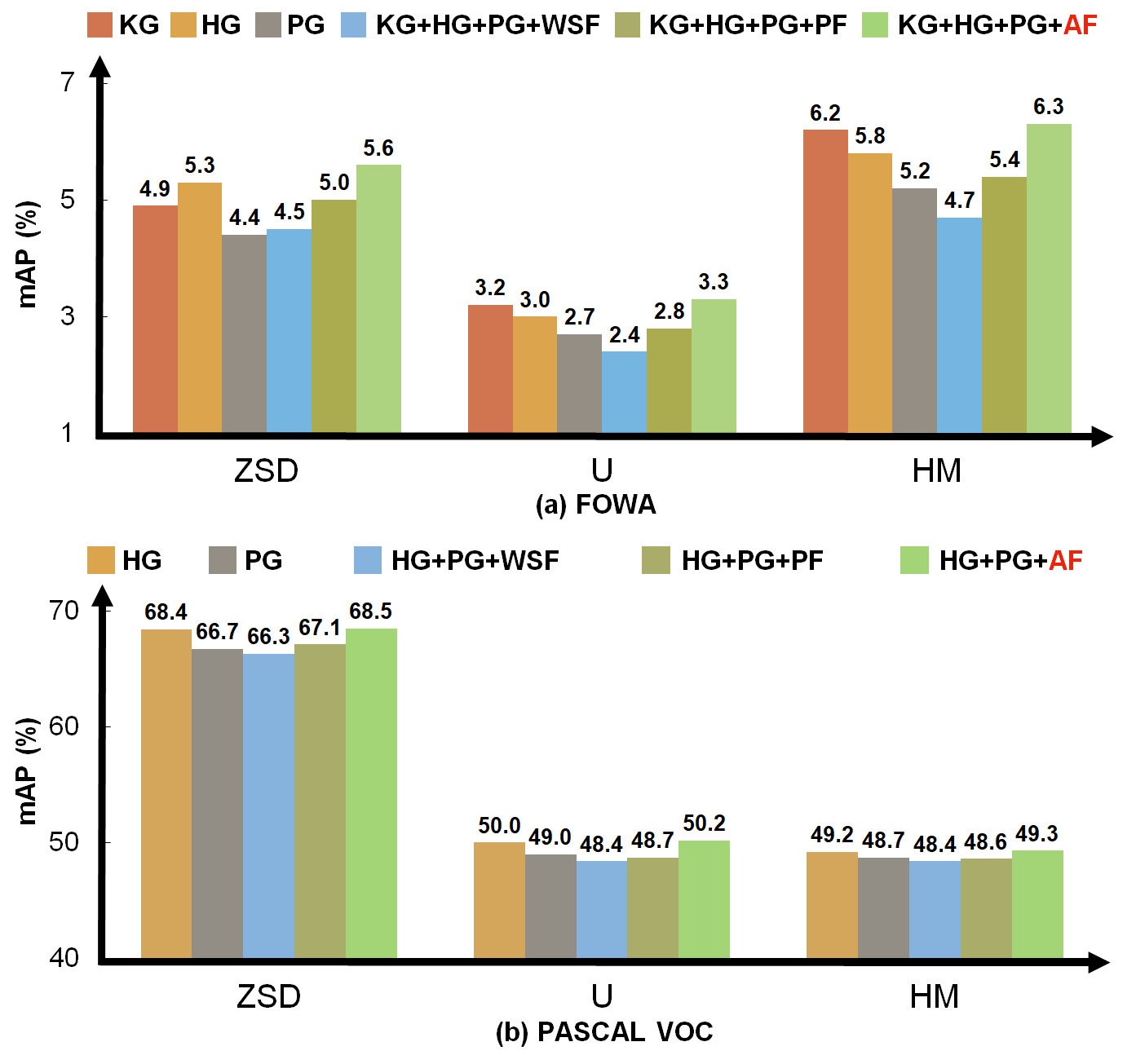}
	\caption{Ablation studies on multi-source graphs and different fusion strategies on FOWA and PASCAL VOC (\%).}
	\label{fig:graph1}
	% \vspace{-0.05cm}
\end{figure}

\begin{figure}[t]
	\centering
	\includegraphics[width=8cm]{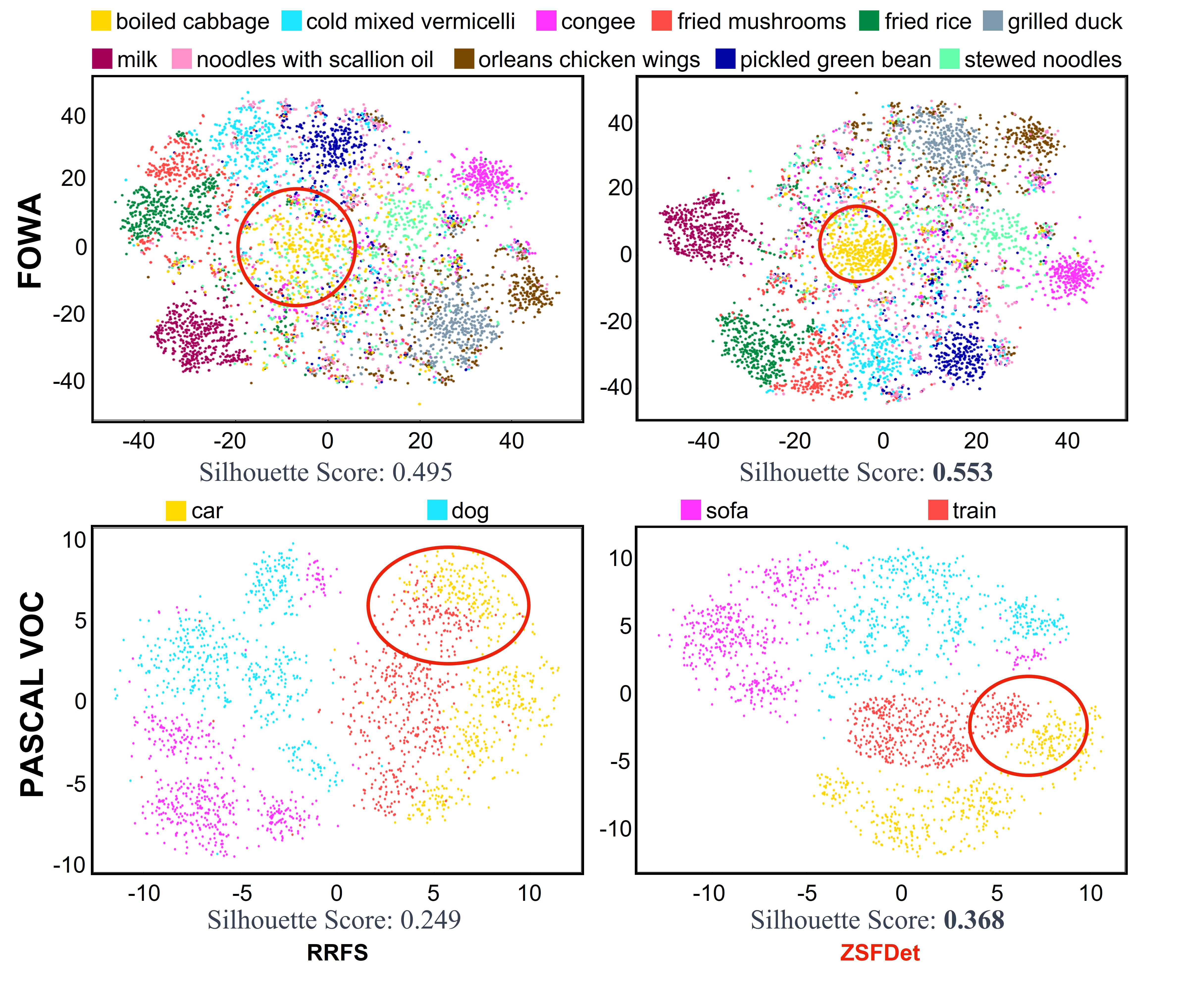}
	\caption{The t-SNE visualization of synthesized features.}
	\label{fig:label_tsne}
	% \vspace{-0.25cm}
\end{figure}

\begin{figure*}[t]
	\centering
	\includegraphics[width=0.99\textwidth]{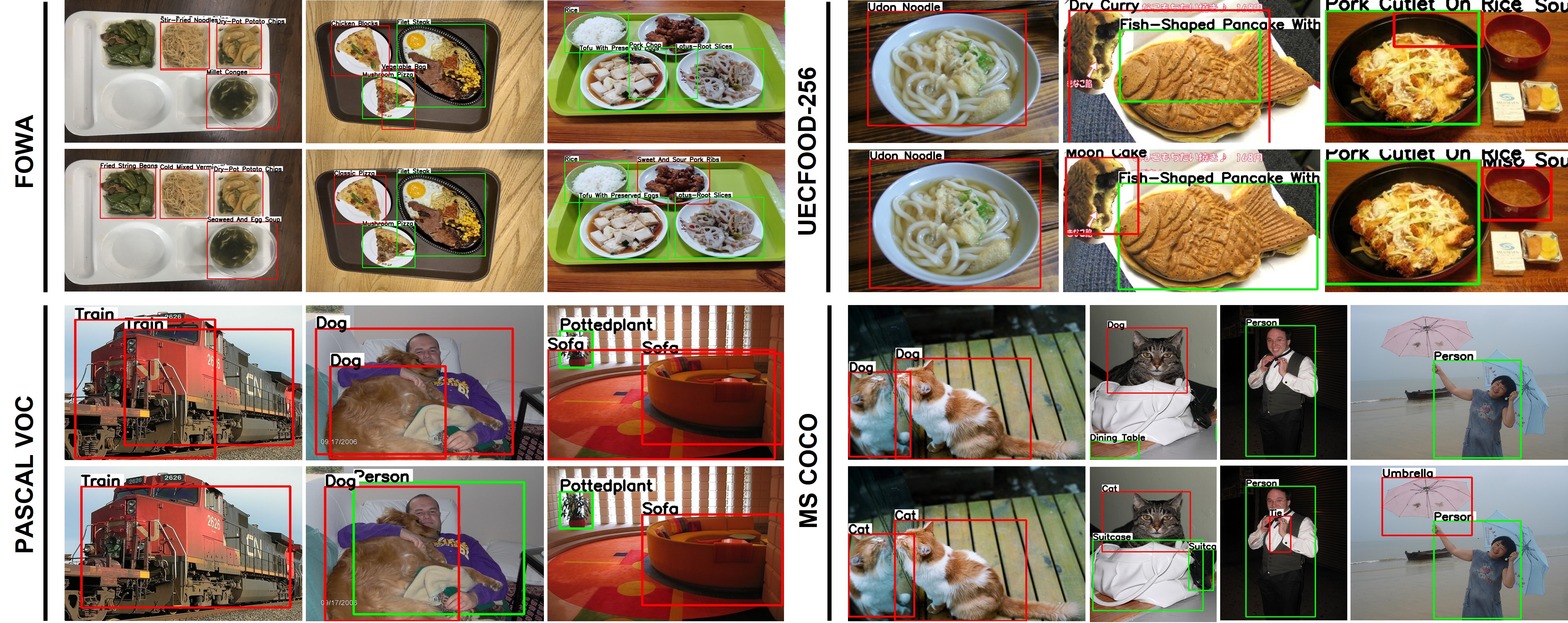}
	\caption{Detection results by baseline RRFS (first row) and our approach (second row) on FOWA, UECFOOD-256, PASCAL VOC and MS COCO. Seen classes are shown with green and unseen with red. Zoom in for a better experience.}
	% \vspace{-0.25cm}
	\label{fig:qualitative}
\end{figure*}

\begin{figure}[!t]
	\begin{center}
		\includegraphics[width=9cm]{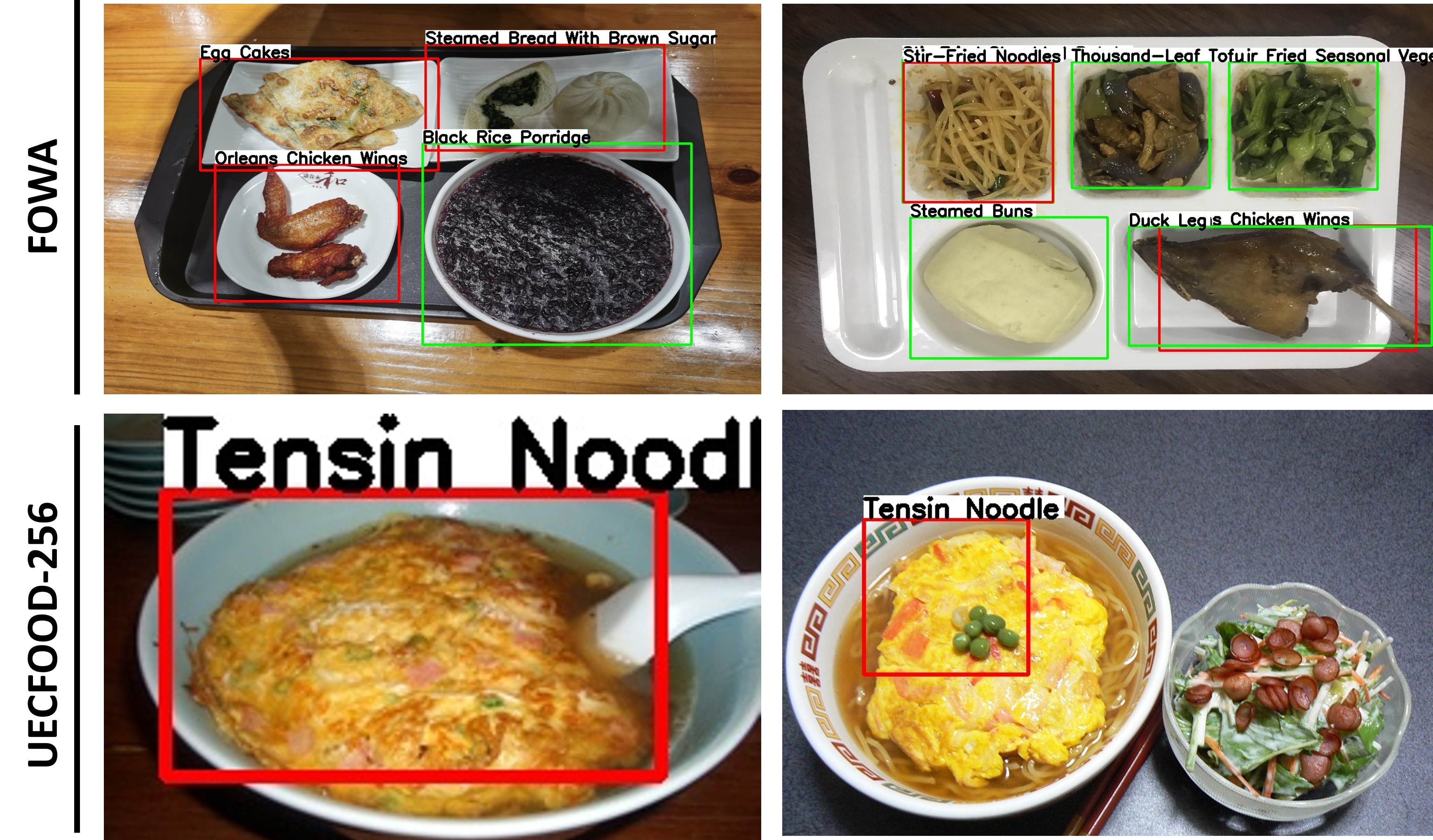}
		\caption{More ZSFD visualization results for failure case studies on the FOWA and UEC-FOOD256.}
		\label{error}
	\end{center}
\end{figure}

\noindent\textbf{Contribution of different modules.}
We first conduct quantitative ablation analysis for the key modules, including the combination of attributes, MSGF, and RFDM. Table \ref{tab:ablation} reports the ``ZSD'' and ``GZSD'' performance of mAP at IoU=0.5 on FOWA and PASCAL VOC. As shown in Table \ref{tab:ablation}, our ablation study reveals that incorporating both word vectors and attribute vectors (``Att.'') enhances the ZSD by 0.2\% on FOWA and 0.3\% on PASCAL VOC, demonstrating the synthesizer's improved ability when leverages richer semantic information in zero-shot scenarios. We can observe that the ``ZSD'' performance has been improved by 1.3\% on FOWA and 2.7\% on PASCAL VOC, and the ``U'' performance has been improved by 0.6\% on FOWA and 1.0\% on PASCAL VOC compared with the baseline when MSGF is implemented. These improvements underscore the MSGF's effectiveness in enhancing feature synthesis and its adaptability across diverse dataset characteristics. Moreover, using RFDM to replace GAN as the core generator further improves mAP 0.5\% of ``ZSD'' performance on FOWA and 0.7\% on PASCAL VOC. It shows that a more powerful generator is crucial for the success of generative zero-shot framework. However, the ``S'' performance in GZSD has not been improved since classifier parameters for seen classes are mainly controlled by the backbone detector.

\noindent\textbf{Contribution of different graphs and fusion strategies.}
Figure~\ref{fig:graph1} compares the performance using three different graphs and fusion strategies in MSGF. We conduct this ablation study based on the GAN as the core generator to eliminate the effects of RFDM. The ``HM'' is improved to 6.2\% mAP by the Knowledge Graph (KG) on FOWA, which is a notable increase compared to Hyperclass Graph (HG) and Probability Graph (PG) strategies. The gained performance by KG suggests its structured representation of domain knowledge benefits the inter-class relationships and attribute correlations modeling that are crucial for ZSFD. The knowledge graph in PASCAL VOC is not designed due to the inaccessibility of the related knowledge graph data. Moreover, various fusion strategies are compared in Figure~\ref{fig:graph1}. The Weighted Sum Fusion (WSF) shows lesser performance improvements, as indicated by the 4.7\% mAP on FOWA and 48.4\% mAP on PASCAL VOC. In contrast, the Attention-based Fusion (AF) strategy significantly outperforms WSF and Product Fusion (PF) strategies, achieving 6.3\% mAP on FOWA and 54.0\% mAP on PASCAL VOC. This performance gain can be attributed to AF's ability to dynamically weigh different semantic sources, thereby providing a more adaptive and discriminative feature synthesis mechanism.

% \vspace{-0.05cm}
\subsection{Qualitative Results}
\label{sec:qualitative}

\noindent\textbf{Feature distribution visualization.}
To further demonstrate the effectiveness of our model in optimizing the distribution structure of visual features, we utilize t-SNE \cite{van2008visualizing} to visualize synthesized unseen features on FOWA and PASCAL VOC in Figure \ref{fig:label_tsne}, where a quarter of the categories on FOWA are randomly selected to make visualization more clear and intuitive. The synthesized features from our model, as compared to the baseline RRFS, show well-defined clusters, as indicated by the silhouette scores: 0.553 versus 0.495 on FOWA, and 0.368 versus 0.249 on PASCAL VOC, respectively. The higher scores, especially for the FOWA dataset, suggest that our approach has resulted in more distinctive features that form well-separated clusters for different classes, facilitating the learning of a robust unseen classifier for ZSD.

% \vspace{-0.05cm}
\noindent\textbf{Detection Results.}
We visualize ZSFD results on FOWA and UECFOOD-256 in Figure~\ref{fig:qualitative}. For each dataset, the first row is detection results by baseline RRFS, and the second row is detection results by our method. As seen in Figure \ref{fig:qualitative}, the baseline method RRFS fails to detect several unseen food objects precisely, while our model provides more robust detection results. Due to the fine-grained problems of food, it is challenging to recognize various food categories, especially for visually similar food objects. 
The introduction of the knowledge graph in ZSFDet can utilize the correlation of ingredients for training an unseen classifier on inter-class separable synthesized features. For example, in the first column on FOWA, ZSFDet can discriminate the \emph{Millet Congee} and the \emph{Seaweed And Egg Soup} that both belong to \emph{Soup} hyperclass via the difference of ingredients. Figure \ref{fig:qualitative} also illustrates that the baseline method is confronted with the mistaken localization (e.g., the \emph{Sofa} in the third sub-figure of the third row), the mistaken classification (the \emph{Dog} in the second sub-figure of the fifth row), and the problem of low recall (the undetected \emph{Umbrella} in the forth sub-figure of the fifth row). Compared with these, our approach also maintains accuracy in detecting general objects. 

We provide more qualitative visualization results for failure case analysis in Figure~\ref{error}. On FOWA, ZSFDet incorrectly identifies \emph{Steamed Bun With Vegetable Stuffing} as \emph{Steamed Bread with Brown Sugar}. This error can arise from the limited visual cues—the stuffing that differentiates the two is barely visible—and their similar semantic attributes. ZSFDet also tends to create multiple bounding boxes for the same object. Addressing this could involve tuning the Non-Maximum Suppression (NMS) parameters or integrating an end-to-end architecture with Hungarian loss. Interestingly, on the UECFOOD-256 dataset, we observed that similar-looking, non-target background objects can disrupt the detection of actual food objects. For example, too much deviation from the ground truth can be observed from the second case on UECFOOD-256 compared with the left correct detection on \emph{Tensin Noodle}. This effect is likely due to the interference from background noise or mislabeling in the scene, which makes fine regression of the target box more difficult, resulting in a lower Intersection over Union (IoU) and greater positional error.

% \vspace{-0.05cm}
\subsection{Discussion}
To further provide a clear understanding of the framework's efficiency in real-world scenarios, we conduct model complexity analysis in TABLE~\ref{complexity} following the rules as defined by~\cite{narayanan2021efficient}. The TABLE~\ref{complexity} compares the complexity of three methods RRF5, SeeDS, and ZSFDet on the FOWA dataset. With respect to the space complexity, the backbone detector contributes most parameters. ZSFDet has a slightly higher computational cost with 658MB and 145.38 GFLOPs, but it surpasses their performance across all metrics. The increased model size and FLOPs ensure a more robust generator and do not affect the inference efficiency. With respect to the time complexity, the training and test speed are barely affected as parallel computing benefits from the added modules. Compared with baselines, despite the close computational costs between SeeDS and ZSFDet, the latter's enhanced ability to detect unseen objects offers significant advantages in practical applications.

Our method achieves state-of-the-art performance on FOWA, UECFOOD-256, and two ZSD datasets by leveraging synthesized robust features from word vectors and attribute vectors. The KEFS in ZSFDet, through MSGF and RFDM, aligns the distribution of synthesized features with real visual features based on the learned knowledge from multi-source graphs and achieves more accurate ZSD detectors on fine-grained classes. While ZSFDet excels in ZSFD, it also generalizes effectively to general ZSD, thanks to probability and hyperclass graphs encoded by multi-source graph mapping and self-attention mechanisms in MSGF. However, the computational increase may be challenging in resource-constrained environments, and attribute reliance might not be viable where data is limited or privacy-sensitive. Performance gains are less pronounced on common datasets like PASCAL VOC and MS COCO, possibly due to the lack of ingredient correlation in these datasets. Despite these challenges, the cross-attention fusion in our knowledge graph induces notable improvements. Future research should aim to mitigate these limitations, exploring how these methods can be simplified for different scenarios without performance loss.

\begin{table}[!t]
	\renewcommand{\arraystretch}{1.3}
	\caption{Model complexity analysis of various methods with their mAP (\%) on FOWA.}
	\label{complexity}
	\centering
	\begin{tabular}{ p{0.8cm}<{\centering}  | p{0.9cm}<{\centering}  p{1.1cm}<{\centering}  p{0.5cm}<{\centering} | p{0.5cm}<{\centering}  p{0.5cm}<{\centering}  p{0.3cm}<{\centering}  p{0.4cm}<{\centering} }
		\hline
		 & Parameter & GFLOPs & FPS & ZSD & S & U & HM \\
		\hline
		RRFS  & 522MB & 117.10 & 15.2 & 4.3   & 82.7  & 2.7   & 5.2  \\
		\hline
            SeeDS  & 652MB & 144.81 & 15.1  & 4.3   & 82.7  & 2.7   & 5.2 \\
            \hline
		ZSFDet & 658MB & 145.38 & 15.1  & 6.1   & 82.8  & 3.6     & 6.9\\
		\hline
	\end{tabular}
\end{table}

%% file: 7_conclusion.tex
% \vspace{-0.15cm}
\section{Conclusion}
\label{sec:conclusion}

This paper benchmarks the Zero-Shot Food Detection (ZSFD) task with fine-grained food attribute annotations in FOWA, which are extracted from the food knowledge graph. We also propose a novel method ZSFDet based on Knowledge-Enhanced Feature Synthesizer (KEFS), which includes Multi-Source Graph Fusion (MSGF) and Region Feature Diffusion Model (RFDM). MSGF enhances feature synthesis based on the rich semantic information from multi-source graph embeddings. RFDM ensures synthesized unseen instance features are diversified and robust for learning the efficient zero-shot detector. Without bells and whistles, the proposed ZSFDet achieves state-of-the-art performance on two ZSFD datasets FOWA and UECFOOD-256, and also outperforms strong baselines on ZSD datasets PASCAL VOC and MS COCO.
Despite its effectiveness, the current computational and data demands highlight areas for improvement. Future directions include optimizing these aspects and integrating the latest large language models, potentially enhancing semantic analysis within an end-to-end Transformer-based detector framework for more reliable performance. Based on powerful vision-language pre-trained models, we envision open-vocabulary food detection becoming increasingly vital in addressing real-world challenges, from dietary monitoring to culinary exploration.